%% file: main.tex
\documentclass[final]{cvpr}

\pdfoutput=1

\usepackage{times}
\usepackage{epsfig}
\usepackage{graphicx}
\usepackage{amsmath}
\usepackage{amssymb}
\usepackage{caption}
\usepackage{subcaption}
\usepackage{threeparttable}

\let\svthefootnote\thefootnote
\newcommand\blankfootnote[1]{%
  \let\thefootnote\relax\footnotetext{#1}%
  \let\thefootnote\svthefootnote%
}

\usepackage[pagebackref=true,breaklinks=true,colorlinks,bookmarks=false]{hyperref}

\def\cvprPaperID{6665} 
\def\confYear{CVPR 2021}

\begin{document}

\title{Panoptic-PolarNet: Proposal-free LiDAR Point Cloud Panoptic Segmentation}

\author{Zixiang~Zhou$^\ast$, Yang~Zhang$^\ast$$^\dagger$, Hassan~Foroosh\\
Department of Computer Science, University of Central Florida\\
{\tt\small \{zhouzixiang, yangzhang\}@knights.ucf.edu, Hassan.Foroosh@ucf.edu}
}

\maketitle


\blankfootnote{$\ast$ Contributed equally.}
\blankfootnote{$\dagger$ Now at Waymo LLC.}
\blankfootnote{Code at: \href{https://github.com/edwardzhou130/Panoptic-PolarNet}{https://github.com/edwardzhou130/Panoptic-PolarNet}.}


\begin{abstract}
Panoptic segmentation presents a new challenge in exploiting the merits of both detection and segmentation, with the aim of unifying instance segmentation and semantic segmentation in a single framework. However, an efficient solution for panoptic segmentation in the emerging domain of LiDAR point cloud is still an open research problem and is very much under-explored. In this paper, we present a fast and robust LiDAR point cloud panoptic segmentation framework, referred to as \textbf{Panoptic-PolarNet}. We learn both semantic segmentation and class-agnostic instance clustering in a single inference network using a polar Bird's Eye View (BEV) representation, enabling us to circumvent the issue of occlusion among instances in urban street scenes. To improve our network's learnability, we also propose an adapted instance augmentation technique and a novel adversarial point cloud pruning method. Our experiments show that Panoptic-PolarNet outperforms the baseline methods on SemanticKITTI and nuScenes datasets with an almost real-time inference speed. Panoptic-PolarNet achieved 54.1\% PQ in the public SemanticKITTI panoptic segmentation leaderboard and leading performance for the validation set of nuScenes.

\end{abstract}

\input{intro}
\input{related}
\input{method}
\input{experiment}
\input{conclusion}

\clearpage

{\small
\bibliographystyle{ieee_fullname}
\bibliography{egbib}
}

\clearpage
\input{supp_arxiv}

\end{document}

%% file: intro.tex
\section{Introduction}

\begin{figure}
    \centering
    \includegraphics[width = 1.0\linewidth]{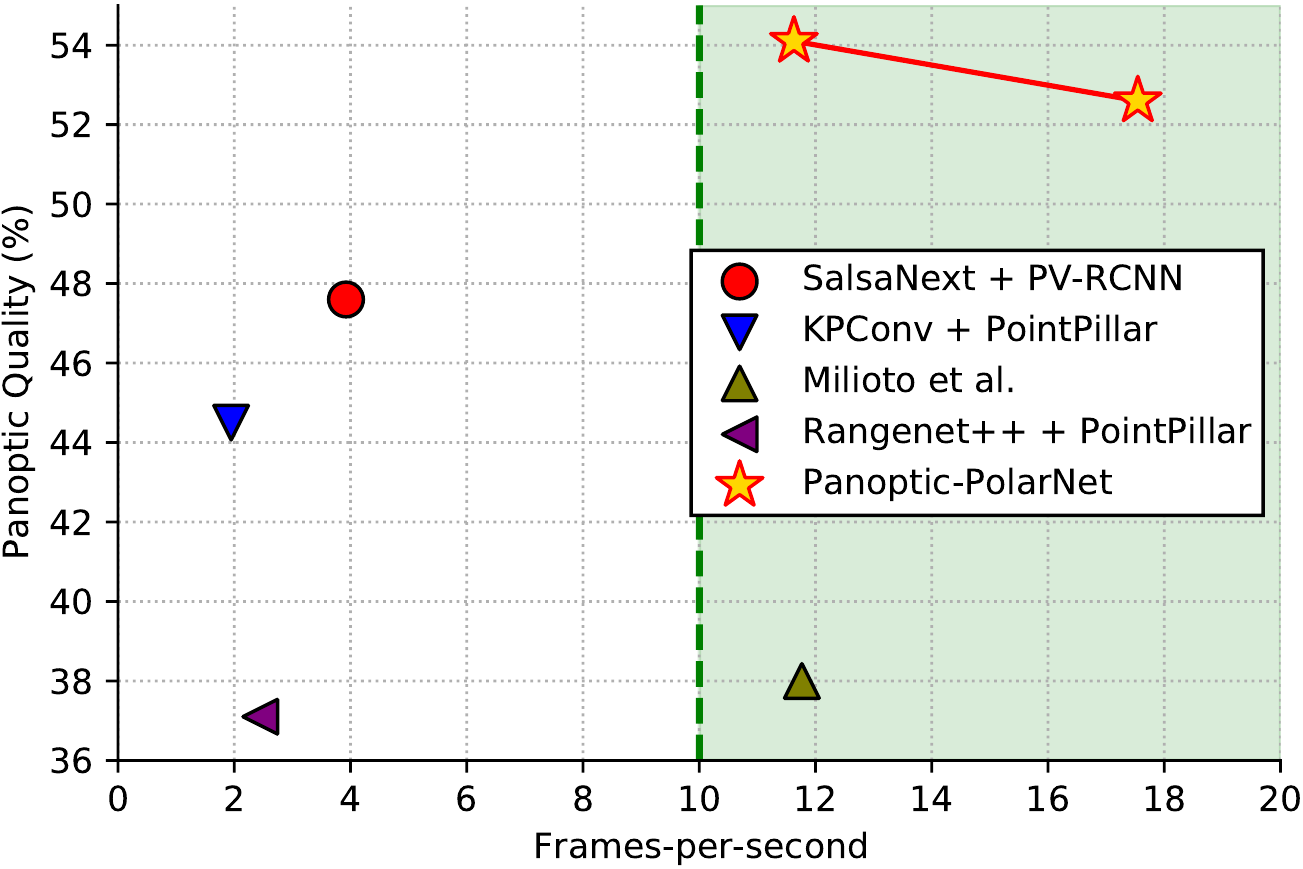}
    \caption{SemanticKITTI~\cite{behley2019iccv} panoptic quality vs. single frame inference latency. The green line marks the sampling rate of the LiDAR scanner, which spins at 10 frames-per-second. Our proposed Panoptic-PolarNet outperforms other baselines in both speed and PQ.}
    \label{fig:teaser}
    \vspace{-20pt}
\end{figure}

As a crucial step in applications such as autonomous driving and robotics, processing and analyzing 3D scanning data have received increasing attention in recent years in computer vision and deep learning. Panoptic segmentation is a recently introduced problem in the image domain~\cite{Kirillov_2019_CVPR} that presents a new challenge in unifying instance segmentation and semantic segmentation in a single training architecture. With the recent introduction of new LiDAR point cloud datasets~\cite{behley2020arxiv,caesar2020nuscenes,geyer2020a2d2} that include both pixel-wise semantic label annotation and object annotation, this problem can now be also explored for 3D scanning data as we propose in this paper.

By definition, Panoptic segmentation requires that we identify both class labels and instance id’s for points in the ``thing’’ classes, and only the class labels for points in the ``stuff’’ classes. To solve this problem, the first question to answer is: What information is needed in order to obtain a panoptic segmentation of data? It can be either the semantic label of all points and the instance clustering of the ``thing’’ classes, or the instance segmentation of the ``thing’’ classes and the class labels of remaining ``stuff’’ classes. As a consequence, these two alternative designs would lead to two different categories of panoptic segmentation, known as proposal-free and proposal-based, the former being adapted from a semantic segmentation network~\cite{long2015fully} and the latter adapted from an object detection network~\cite{he2017mask}.

2D image panoptic segmentation faces two main problems. First, proposal-based ones segment instances independently within each individual object proposal. Such approaches require extra architectural modifications ~\cite{liu2019end,xiong19upsnet} to compensate for the impact of heavy object collision in the proposals. Second, semantic segmentation and instance segmentation are usually handled in two separate prediction heads in order to tailor the design of the dedicated network to each task. However, this may inevitably introduce either potential conflicts or redundant information since these two tasks clearly share common characteristics. For example, in the proposal-based methods, semantic and instance heads can yield different label predictions at the same pixel. And in the proposal-free methods, the features learned in the instance head have significant correlations with class labels. Both cases ultimately lead to inference inefficiency.

3D panoptic segmentation, on the other hand, is by and large at its infancy and still an open research problem. It is mainly motivated by LiDAR point cloud processing in applications such as self-driving cars, autonomous robot navigation, and environment mapping, all of which generally require real-time processing. On the other hand, compared to conventional 3D data in computer vision, LiDAR point clouds are irregularly sampled in the 3D space. These differences in terms of the nature of the 3D data, the need for real-time processing, and the level of accuracy needed for safety and security (e.g., in self-driving cars) are clearly creating new challenges, encouraging new innovative solutions. These challenges motivated us to find a more suitable architecture that takes into account the unique characteristics of LiDAR data, efficiently solves panoptic segmentation with minimum conflicts in predictions (instance versus class), and achieves real-time or near real-time speed without compromising accuracy. 

Given the speed limitation, proposal-free methods naturally seem to be a more favorable choice, since they are proven to perform better in computational time in the 2D case. Therefore, starting from a backbone semantic prediction network~\cite{Zhang_2020_CVPR}, our first goal is to integrate it with a network for class-agnostic instance clustering. We hypothesize that most ``thing'' class objects in the LiDAR point cloud are separable when projected onto the XY-plane. Instance separability implies that the discretized BEV representation~\cite{wang2019pseudo} is highly suitable for LiDAR point cloud instance clustering. Therefore, we can use the same network of PolarNet also to generate discriminative features for separating instances in the BEV. Based on these observations and assumptions, we propose a panoptic segmentation framework that simultaneously learns semantic and instance features on the discretized BEV map. Therefore, we follow the backbone network design of PolarNet~\cite{Zhang_2020_CVPR} to generate the 3D semantic prediction and use a lightweight 2D instance head inspired by Panoptic-DeepLab~\cite{cheng2020panoptic} on top of it. Predictions from semantic and instance heads are then fused through a majority voting to create the final panoptic segmentation. This results in a highly efficient proposal-free panoptic segmentation network design, which we refer to as \textbf{Panoptic-PolarNet}.

We evaluated our approach on SemanticKITTI and nuScenes datasets. Panoptic-PolarNet achieves state-of-the-art performance. Compared to the PolarNet, our instance segmentation head only introduces 0.1M parameters and increases the inference time by only 0.027s. 

Our contributions are summarized as follows:

\begin{itemize}
    \item We propose a model taking into account the specific nature of LiDAR data and the applications in mind, to construct a proposal-free LiDAR panoptic segmentation network that can efficiently cluster instances on top of the semantic segmentation.
    \item Unlike existing panoptic segmentation networks that generally use two entirely separate decoding modules for semantic and instance segmentation and rely on an attention module to connect the learned information, our networks share decoding layers among them, allowing for early fusion at feature extraction level. This early fusion strategy has two substantial impacts: (1) it reduces redundancy between the networks and therefore improves computational efficiency; (2) increases the PQ measure despite a smaller computation load.
    \item Compared to existing proposal-based panoptic segmentation methods that suffer from class and instance prediction overlapping, we propose a proposal-free design and train the instance head without bounding box annotation, which allows us to avoid the conflict of class prediction.
    \item We introduce two novel point cloud data augmentation methods that can apply to any other LiDAR segmentation networks.
    \item Experiments show that our approach outperforms strong baselines on SemanticKITTI and nuScenes datasets with smaller and near-real-time latency, as shown in Figure~\ref{fig:teaser}.
\end{itemize}

%% file: related.tex
\section{Related works}

\subsection{Image based panoptic segmentation}
Current 2D panoptic segmentation methods normally divide panoptic segmentation into two subproblems: semantic segmentation and instance segmentation. They are trained in a single network with a shared feature encoding layer and separated heading layer. According to how they accurately separate different instances, panoptic segmentation methods can be categorized into top-down/ proposal-based and bottom-up/ proposal-free approaches. Top-down methods usually use Mask R-CNN~\cite{he2017mask} to get each object's instance mask first and then fill in the rest region with the semantic segmentation prediction. While this design gives a reliable instance segmentation result, it requires additional means to resolve the overlapping instances and the conflict between instance and semantic predictions. Liu \textit{et al.}~\cite{liu2019end} propose a spatial ranking module to sort the overlapping masks. UPSnet~\cite{xiong19upsnet} introduces a panoptic head to resolve the conflict between instance and semantic predictions by adding an unknown class label. EfficientPS~\cite{mohan2020efficientps} proposes a panoptic fusion module that dynamically adapts the fusion of instance and semantic heads according to their confidence. Recent research also focuses on designing either end-to-end training~\cite{liu2019end,xiong19upsnet} or attention module bridging between semantic and instance learning~\cite{li2019attention,Wu_2020_CVPR,Chen_2020_CVPR}. On the contrary, bottom-up methods generally get semantic prediction and then fuse it with class-agnostic instance segmentation. The first bottom-up method, DeeperLab~\cite{yang2019deeperlab}, proposes to separate the instance using bounding box corners and center. Panoptic-DeepLab~\cite{cheng2020panoptic} further simplifies this grouping method by predicting the instance center and offset. SSAP~\cite{gao2019ssap} uses cascaded graph partition to segment instance from a pixel-pair affinity pyramid.

\subsection{LiDAR point cloud object detection and semantic segmentation}
Compared to the conventional 3D point cloud data, LiDAR point cloud is inherently 2.5D data since it is a perspective projection of the real world. This results in a sparse and imbalanced distribution of points among 3D geometrical space. Besides, most LiDAR point cloud tasks are targeted on the autonomous driving scenario, which creates even larger data size for the conventional point cloud method~\cite{qi2017pointnet,qi2017pointnet++,qi2019deep} to process. In addition to directly learning features on either the point level~\cite{shi2019pointrcnn,thomas2019KPConv,hu2019randla} or voxelized space~\cite{zhou2018voxelnet,yan2018second}, research on LiDAR point cloud also uses projected space like the bird's eye view~\cite{lang2019pointpillars,yang2018pixor,Porzi_2019_CVPR} or spherical projection/range image~\cite{qi2018frustum,wu2018squeezeseg,wu2019squeezesegv2,milioto2019rangenet++,cortinhal2020salsanext}, and sometimes the fusion of multiple aforementioned views~\cite{zhou2020end,tang2020searching}. Like its counterpart in 2D object detection, LiDAR point cloud object detection methods are also divided into proposal-based and proposal-free ones. Proposal-based methods~\cite{zhou2018voxelnet,shi2019pointrcnn,yan2018second} first generate region proposal from an encoded feature and use another head to select and refine object bounding box, while proposal-free methods directly predict object through vote clustering~\cite{qi2019deep} or keypoint/center estimation~\cite{yin2020center}. For the segmentation problem, researchers pay more attention to efficiently extracting and recovering local and global context. KPConv~\cite{thomas2019KPConv} and RandLA~\cite{hu2019randla} proposed to use kernel point convolution and local feature aggregation module to replace the conventional convolutional layer in an encoder-decoder structure to operate on the point cloud directly. However, it requires a time-consuming preprocessing to build the graph. Many other methods~\cite{wu2018squeezeseg,cortinhal2020salsanext} choose to use 2D convolution to segment the point cloud on the 2D point projection. Rangenet++~\cite{milioto2019rangenet++} and KPRNet~\cite{kochanov2020kprnet} introduce additional KNN and aligning processing to better recover label from a projected view to the original point cloud. PolarNet~\cite{Zhang_2020_CVPR} encodes point cloud into a polar BEV to compensate for the imbalanced distribution of points in the physical space.

\subsection{Point cloud panoptic segmentation}
As a rising task, LiDAR panoptic segmentation has not been well studied yet. However, many researchers have already explored indoor point cloud panoptic segmentation by combining instance segmentation and semantic segmentation methods. Most of them~\cite{wang2019associatively,lahoud20193d,pham2019jsis3d,Liu2020SelfPredictionFJ} use the discriminative loss~\cite{de2017semantic} to learn a embedded feature space to cluster instances. Zhou \textit{et al.}\cite{zhou2020joint} extract the instance segmentation from the region proposals clustered from the semantic segmentation. SemanticKITTI~\cite{behley2020arxiv} benchmarks the first panoptic segmentation LiDAR dataset by combining existing state-of-the-art object detection and semantic segmentation networks. MOPT~\cite{hurtado2020mopt} attaches a semantic head to Mask R-CNN  to generate panoptic segmentation on the range image. Milioto \textit{et al.}~\cite{milioto2020iros} proposes to solve LiDAR point cloud panoptic segmentation on the range image first then restore it to point cloud level through a tri-linear upsampling.

%% file: method.tex
\section{Panoptic-PolarNet}

\begin{figure*}
\vspace{-15pt}
\centering
  \includegraphics[width=0.90\textwidth]{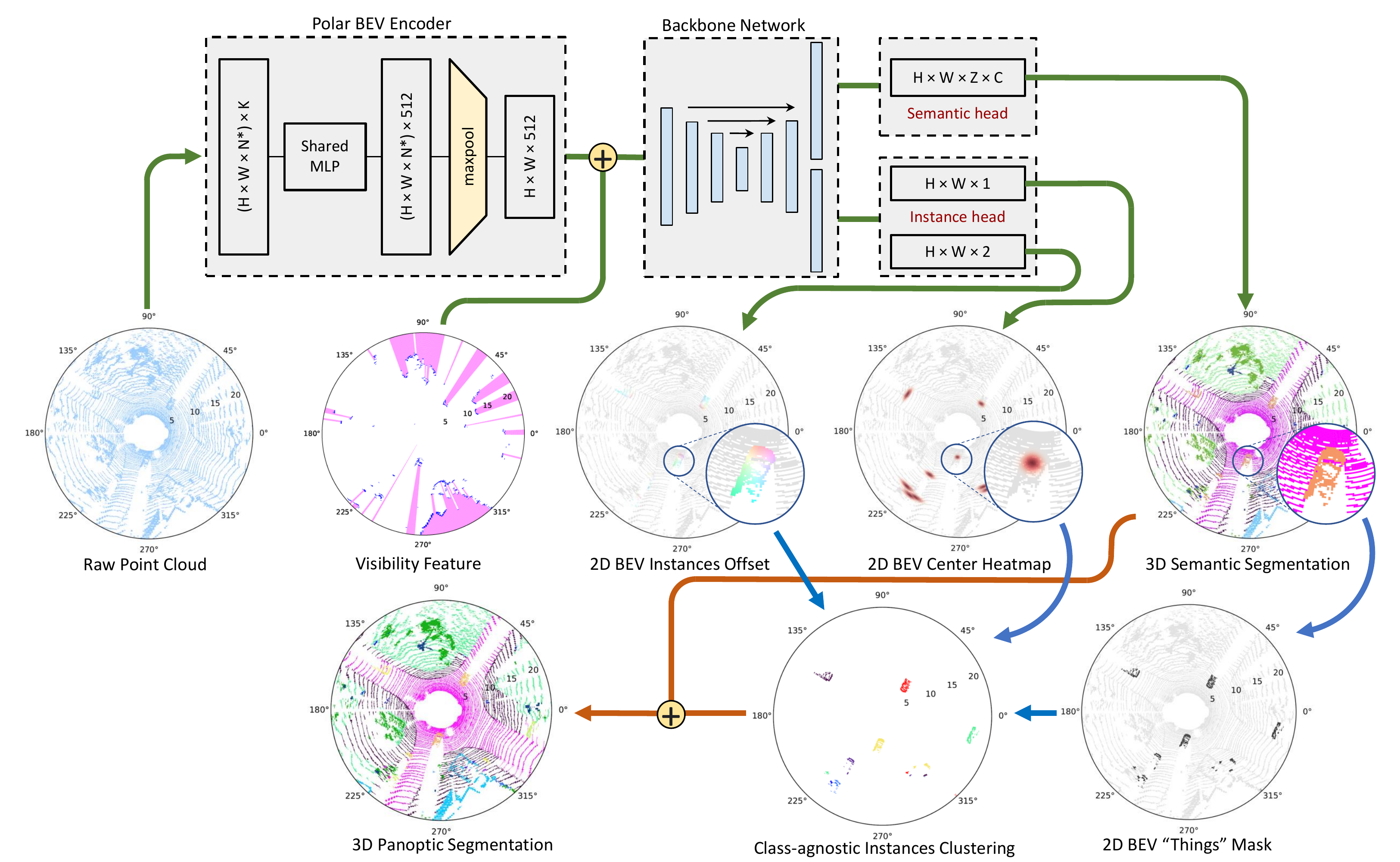}
      \caption{Our Panoptic-PolarNet framework. We first encode the raw point cloud data with $K$ features into a fixed-size representation on the polar BEV map. Next, we use a single backbone encoder-decoder network~\cite{Zhang_2020_CVPR} to generate semantic prediction, center heatmap and offset regression. Finally, we merge these outputs via a voting-based fusion to yield the panoptic segmentation result.}
  \label{fig:architecture}
  \vspace{-15pt}
\end{figure*}

As shown in Figure~\ref{fig:architecture}, our Panoptic-PolarNet consists of the following four components: (1) a network that encodes the raw point cloud data to a fixed-size 2D polar BEV representation, (2) a shared encoder-decoder backbone network, (3) two independent heads for semantic and instance segmentation, (4) a fusion step that merges the aforementioned predictions into one final panoptic segmentation result. 

\subsection{Preliminary}
Given a set of points $P = \{(x,y,z,r)_{n}|n\in\{1,\dots,N\}\}$, where $(x,y,z)$ are the 3D coordinates relative to the LiDAR scanner's reference frame and $r$ is the intensity of reflection, and a set of ground truth class labels $C_{GT}= \{l_{n}|n\in\{1,\dots,N\}\}$, LiDAR point cloud semantic segmentation task aims to predict a unique set of class labels $C_{p}$ for the points $P$ that minimizes the difference with $C_{GT}$. Panoptic segmentation task extends this problem to requiring that points belonging to separate instances have different labels in some ``thing'' classes, e.g., car, bicycle, and human. The remaining classes are ``stuff'' classes, which do not require detailed separation and share the same label among all points.

\subsection{Polar BEV encoder}
To process a point cloud containing a random size of points, we need to create a fixed-size representation through projection and quantization. We use BEV for two main reasons. First, BEV provides a trade-off between computational cost and accuracy, enabling us to use the more efficient 2D convolutional networks to process the data. Second, since objects rarely overlap along the z-axis in the urban scene, BEV is empirically the best projection for object detection~\cite{wang2019pseudo}. We also represent the points in the polar coordinates rather than conventional Cartesian coordinates to balance the distribution of points among different ranges~\cite{Zhang_2020_CVPR}. The polar coordinate gives neural networks better potential to learn discriminative features at locations closer to the sensor and minimizes the information losses due to quantization.

We adopt the original polar BEV encoder design from PolarNet~\cite{Zhang_2020_CVPR}. More specifically, we first group a point cloud data $P\in \mathbb{R}^{N\times K}$ to $P'\in \mathbb{R}^{(H\times W\times N^{\star})\times K}$ based on its position in the polar BEV map, where $K$ is the input feature dimension, $H$ and $W$ are the grid size of the BEV map and $N^{\star}$ is the number of points in each BEV grid. Next, we encode this point cloud through a simplified PointNet~\cite{qi2017pointnet}, which only contains MLP. Then, a max-pooling layer is applied at each BEV grid to create a fixed-size representation $M\in \mathbb{R}^{H\times W\times C}$, where $C$ is the feature channel. We use $C=512$ in our experiment.

\subsection{Semantic Segmentation}
After encoding LiDAR point cloud data into a feature matrix $M$, most 2D semantic segmentation backbone networks are able to process it. We follow PolarNet to use Unet~\cite{ronneberger2015u} with 4 encoding layers and 4 decoding layers as the backbone network. Unlike other panoptic segmentation networks that generally use two entirely separate decoding modules for semantic and instance segmentation, our network shares the first three decoding layers among them. Our semantic head generates multiple predictions at each pixel $C_{p}\in\mathbb{Z}^{Z\times H\times W}$, which are later reshaped back to 3D voxels to separate labels at different heights along $Z$-axis. We calculate the loss at the voxel level during the training, where the groundtruth label for each voxel is decided by majority voting of points within the same voxel.

\subsection{Panoptic Segmentation}
According to \cite{Kirillov_2019_CVPR}, one big problem in 2D image panoptic segmentation~ is the difficulty to efficiently separate instances when the collision occurs, e.g., two people standing next to each other. We hypothesized that we could circumvent this challenge in LiDAR data based on two assumptions. First,  objects rarely collide in 3D space even their masks overlap in 2D projection.  Second, most ``thing'' class objects in the LiDAR point cloud of urban scenes are still separable when projected onto the XY-plane from 3D space. Such a claim is also supported by \cite{wang2019pseudo}, who find the same object detection network has better performance in BEV in contrast to 2D projection. This suggests that the BEV representation has the potential to not only improve the performance but also reduce the problem of instance clustering in the LiDAR point cloud panoptic segmentation to a 2D problem. Therefore, we can use the same network of PolarNet to generate discriminative features for separating instances in the BEV.

We follow the instance head design in Panoptic-DeepLab~\cite{cheng2020panoptic} to predict the center heatmap and the offset to the object center for each BEV pixel. Pixels that have the same nearest center are grouped together. Compared to other top-down methods with overlaps of class prediction between segmentation and instance branches, this bottom-up design provides only class-agnostic instance grouping. This allows us to avoid the conflict of class prediction and train instance head without bounding box annotation. During the training phase, we encode the ground truth center map by a 2D Gaussian distribution around each instance's mass center. For each pixel $p$ in the BEV map, the heatmap is $H_{p} = \max_i\exp(-\frac{(p-c_{i})^{2}}{2\sigma^{2}})$, where $c_{i}$ is the mass center of one instance in the polar BEV coordinates.

To merge the 3D semantic segmentation and 2D instance grouping predictions, we propose a fusion step as shown in Figure~\ref{fig:architecture}. First, the top $k$ centers are selected from the heatmap prediction after a non-maximum suppression. Next, we use the semantic segmentation prediction to create a foreground mask where at least one ``thing'' class is detected at one BEV pixel. Pixels in the foreground are then grouped together based on the minimal distance $d(p,c_{i}) = \parallel p +  \mbox{offset}(p)-c_{i}\parallel_{2}$ to one of the $k$ centers. Lastly, ``thing'' class predictions in the semantic segmentation head are assigned a unique instance label $L$ for each group $G_{i}$ in the BEV using a majority voting according to semantic segmentation probability $P(v)$: $L_{i} = \mathrm{argmax}\sum_{v\in G_{i}} P(v)$. All these operations are implemented in GPU, requiring little computational time.

\subsection{Augmenting Panoptic-PolarNet}

\textbf{Instance augmentation:} Training data augmentation on the instance level has proven to be an important technique for LiDAR object detection~\cite{lang2019pointpillars,hahner2020quantifying} without increasing inference computational cost. How the sensor samples points of an instance is determined by the sensor's angle interval, the relative pose, and distance of an instance to the sensor. Our instance augmentation aims to increase the variance of data without changing the projection properties of instance points. We summarise it as the following three steps: (1) Instance oversampling: We randomly choose 5 instances from the whole training set and paste them into the current training scan. The probability of each class being selected is in proportion to the reciprocal of its point distribution ratio. The imported points retain the same relative coordinates and reflection values as in their source. (2) Instance global augmentation: The goal here is to find a transformation to change an instance's location on XY-plane without altering its projection on the sensor. The need to preserve projection narrows the transformation to either rotation on the center or reflection on a certain view plane through the center. We apply those two transformations to each instance with a 20\% probability for each transformation. (3) Instance local augmentation: We also apply small independent translation and rotation to each instance, which serves as measurement noise. We sample the translations $[\Delta x,\Delta y,\Delta z]$ from a normal distribution $N(0,0.25)$, and the rotation angle $\Delta\theta$ from a uniform distribution $U(-20/\pi,+20/\pi)$.

\textbf{Point cloud self-adversarial pruning:} Inspired by YOLO-v4~\cite{bochkovskiy2020yolov4}, we also use self-adversarial pruning on the point cloud after the training is almost converged. The idea of self-adversarial pruning is to find the most influential points through the network itself. Those points are likely to be either noise or key feature points. By omitting those points during training, we enforce the network to learn more general features from the overall point cloud instead of overfitting into some specific geometry patterns. More specifically, we use two forward-backward loops for each batch of input data. We use the gradient to select those highly influential points in the first forward-backward loop and feed the altered data to the second forward-backward loop to update the network weights after omitting those points. Similar to \cite{molchanov2019importance}, we consider the gradient variance as the diagonal of the Fisher information matrix, which represents the importance of input to the panoptic loss. In the experiments, we deleted only the top 1\% of the points according to validation results. 

\textbf{Visibility feature:} Visibility is a concept commonly used in the mapping problem to create an occupancy map of the environment through raycasting. Recently, Hu \textit{et al.}~\cite{Hu_2020_CVPR} included the visibility in the detection problem to enrich the voxel representation of the point cloud. Given a point $(x,y,z)$ in the LiDAR point cloud, the space along the same direction $\alpha(x,y,z)$ can be divided into visible if $0<\alpha<1$ and occluded if $\alpha>1$. However, computing the visibility for the whole 3D space requires traversing through all points, which is usually preprocessed offline before the training. Since the range at the $z$ axis is much smaller than the other two axes in the LiDAR point cloud, we approximate this traversal as for each point at $(d,\theta,z)$, where $d$ and $\theta$ are the distance and angle in the polar coordinates, the space of $(\alpha d,\theta,z)$ is visible if $0<\alpha<1$ and occluded if $\alpha>1$. Hence we can compute the visibility for each voxel efficiently in the polar coordinate alongside the instance data augmentation during the training. We concatenate the visibility feature with the feature representation generated by the polar BEV encoder, then feed it into the backbone network in our implementation.

%% file: experiment.tex
\section{Experiment}
In this section, we demonstrate our panoptic segmentation results on the SemanticKITTI~\cite{behley2020arxiv} dataset and the nuScenes~\cite{caesar2020nuscenes} dataset. Due to page limitations, please refer to our supplementary material for more details on the experiments, discussions, and qualitative examples.

\begin{table*}
\centering
\caption{Panoptic Segmentation results on the \textbf{test} split of SemanticKITTI.}
\vspace{-10pt}
\label{tab:panoptic_test_semantickitti}
\resizebox{\linewidth}{!}{
\begin{threeparttable}
\begin{tabular}{|l|c|cccc|ccc|ccc|c|}
\hline
Method                     & Latency  & PQ     & PQ$^{\dagger}$     & RQ     & SQ     & PQ$^{Th}$     & RQ$^{Th}$     & SQ$^{Th}$     & PQ$^{St}$     & RQ$^{St}$     & SQ$^{St}$  & mIoU\\ \hline
Rangenet++~\cite{milioto2019rangenet++}  +  PointPillar~\cite{lang2019pointpillars} & 0.409s   & 37.1\% & 45.9\% & 47.0\% & 75.9\% & 20.2\% & 25.2\% & 75.2\% & 49.3\% & 62.8\% & 76.5\% & 52.4\%\\
Milioto \textit{et al.}~\cite{milioto2020iros} & 0.085s   & 38.0\% & 47.0\% & 48.2\% & 76.5\% & 25.6\% & 31.8\% & 76.8\% & 47.1\% & 60.1\% & 76.2\% & 50.9\%\\
KPConv~\cite{thomas2019KPConv}  +  PointPillar~\cite{lang2019pointpillars}     & 0.514s   & 44.5\% & 52.5\% & 54.4\% & 80.0\% & 32.7\% & 38.7\% & 81.5\% & 53.1\% & 65.9\% & \textbf{79.0\%} & 58.8\%\\
SalsaNext~\cite{cortinhal2020salsanext} + PV-RCNN~\cite{shi2020pv}        & 0.255s   & 47.6\% & 55.3\% & 58.6\% & 79.5\% & 39.1\% & 45.9\% & 82.3\% & 53.7\% & 67.9\% & 77.5\% & 58.9\%\\
Panoster~\cite{gasperini2020panoster} & -\tnote{*}   & 52.7\% & 59.9\% & 64.1\% & 80.7\% & 49.9\% & 58.8\% & 83.3\% & \textbf{55.1\%} & \textbf{68.2\%} & 78.8\% & \textbf{59.9\%}\\
\hline
Panoptic-PolarNet-mini          & 0.057s   & 52.6\% & 59.4\% & 63.6\% & 80.9\% & 51.9\% & 59.5\% & 86.9\% & 53.1\% & 66.6\% & 76.5\% & 58.4\%\\ 
Panoptic-PolarNet          & 0.086s   & \textbf{54.1\%} & \textbf{60.7\%} & \textbf{65.0\%} & \textbf{81.4\%} & \textbf{53.3\%} & \textbf{60.6\%} & \textbf{87.2\%} & 54.8\% & 68.1\% & 77.2\% & 59.5\%\\ \hline
\end{tabular}
\begin{tablenotes}
   \item[*] ~\cite{gasperini2020panoster} did not disclose their latency nor did they release their code. Since \cite{gasperini2020panoster} is a variant of KPConv, its latency should be similar to KPConv which is stated to be 200ms~\cite{thomas2019KPConv}.
 \end{tablenotes}
\end{threeparttable}
}
\vspace{-10pt}
\end{table*}

\begin{table*}
\centering
\caption{Panoptic Segmentation results on the \textbf{validation} split of SemanticKITTI.}
\vspace{-10pt}
\label{tab:panoptic_val_semantickitti}
\resizebox{\linewidth}{!}{
\begin{tabular}{|l|c|cccc|ccc|ccc|c|}
\hline
Method                     & Latency    & PQ & PQ$^{\dagger}$     & RQ     & SQ     & PQ$^{Th}$     & RQ$^{Th}$     & SQ$^{Th}$     & PQ$^{St}$     & RQ$^{St}$     & SQ$^{St}$     & mIoU\\ \hline
MOPT~\cite{hurtado2020mopt} & 0.146s   & 40.0\% & - & 48.3\% & 73.0\% & 29.9\% & 33.6\% & 76.8\% & 47.4\% & \textbf{70.3\%} & 59.1\% & 53.8\%\\
SalsaNext~\cite{cortinhal2020salsanext} + PointRCNN~\cite{shi2019pointrcnn}        & 0.196s   & 47.5\% & 53.2\% & 58.2\% & 74.4\% & 42.5\% & 50.3\% & 73.4\% & 51.1\% & 64.0\% & \textbf{75.2\%} & 59.0\%\\
PolarNet~\cite{Zhang_2020_CVPR} + PointRCNN~\cite{shi2019pointrcnn}        & 0.202s   & 47.9\% & 53.1\% & 58.7\% & 71.3\% & 39.5\% & 47.8\% & 71.0\% & 54.1\% & 66.7\% & 71.4\% & 58.2\%\\
PolarNet~\cite{Zhang_2020_CVPR} + PV-RCNN~\cite{shi2020pv}        & 0.261s   & 49.9\% & 55.0\% & 60.9\% & 71.4\% & 44.1\% & 52.9\% & 71.2\% & 54.1\% & 66.7\% & 71.4\% & 58.2\%\\
SalsaNext~\cite{cortinhal2020salsanext} + PV-RCNN~\cite{shi2020pv}        & 0.255s  & 49.9\% & 55.6\% & 61.0\% & 74.4\% & 48.3\% & 56.7\% & 73.3\% & 51.1\% & 64.0\% & \textbf{75.2\%} & 59.0\% \\ \hline
Panoptic-PolarNet in Cartesian coordinates          & 0.078s   & 54.3\% & 58.8\% & 65.5\% & 78.0\% & 58.4\% & 67.3\% & 85.3\% & 50.3\% & 64.2\% & 69.1\% & 58.6\%\\
Panoptic-PolarNet-mini          & 0.057s   & 57.1\% & 61.8\% & 68.1\% & 77.7\% & 63.8\% & 72.6\% & \textbf{87.4\%} & 52.2\% & 64.9\% & 70.6\% & 61.9\%\\ 
Panoptic-PolarNet          & 0.086s   & \textbf{59.1\%} & \textbf{64.1\%} & \textbf{70.2\%} & \textbf{78.3\%} & \textbf{65.7\%} & \textbf{74.7\%} & \textbf{87.4\%} & \textbf{54.3\%} & 66.9\% & 71.6\% & \textbf{64.5\%}\\ \hline
\end{tabular}
}
\vspace{-10pt}
\end{table*}

\subsection{Datasets}

\textbf{SemanticKITTI} provides point-wise semantic and instance annotations for the well known KITTI~\cite{geiger2012cvpr} odometry dataset, which contains 10/1/11 training/validation/testing sequences, and a total of 43551 LiDAR scans of European city streets. Each SemanticKITTI scan has 104452 points on average and is annotated with 20 class labels, 8 of which are selected as ``thing'' classes.

\textbf{NuScenes} is a large scale autonomous driving dataset created by Motional. It contains 1000 driving scenes, with 850 scenes for training and validation, and 150 scenes for testing. In each keyframe that is sampled every 0.5s, nuScenes provides bounding box annotations for 10 possible object classes and point-wise semantic labels for 16 classes. The first 10 are the same as object classes. Although nuScenes also provides image and radar data, we only used the LiDAR data in the keyframe during the training and validation. Unlike SemanticKITTI, nuScenes does not explicitly provide the instance label for each point.  We manually created the panoptic instance annotation by assigning ``thing'' points to its closest detection bounding box. We remove outliers by omitting the ``thing'' points that are more than 5m apart from the nearest bounding box centroids. Since nuScenes does not provide panoptic segmentation metrics on the test set evaluation server, we only report panoptic segmentation results on the validation set.

\subsection{Baselines}

Our baselines include both dedicated panoptic LiDAR point cloud segmentation methods as well as combinations of state-of-the-art segmentation and detection pipelines. The method proposed by Milioto \textit{et al.}~\cite{milioto2020iros}, MOPT~\cite{hurtado2020mopt} and Panoster~\cite{gasperini2020panoster} are the only three methods specifically designed for LiDAR point cloud panoptic segmentation. The first two are trained on range images, while the third one is a variant of KPConv~\cite{thomas2019KPConv} at the point level. Due to the lack of public implementation, we use their reported results for comparison. For the combining methods, we pick the highest-ranking approaches with public implementation. We use PolarNet~\cite{Zhang_2020_CVPR} and SalsaNext~\cite{cortinhal2020salsanext} to generate the semantic prediction. We use PV-RCNN~\cite{shi2020pv} and PointRCNN~\cite{shi2019pointrcnn} to generate the object bounding box prediction for the SemanticKITTI dataset. In addition, we include two combining baselines (Rangenet++~\cite{milioto2019rangenet++}/KPConv~\cite{thomas2019KPConv} + PointPillar~\cite{lang2019pointpillars}) from~\cite{behley2020arxiv} in our comparison.  We generate all results of these baselines from their publicly available implementations and pretrained-networks. During the combination, we pick the points within and close to each bounding box prediction and assign a unique instance to all points that have the same class as the bounding box. We use the combined time of semantic segmentation and object detection as the total inference time for the combining method. On the nuScenes dataset, we use OpenPCDet's~\cite{openpcdet2020} pre-trained nuScenes CBGS~\cite{zhu2019class} model for the object detection. Since SalsaNext~\cite{cortinhal2020salsanext} has the best SemanticKITTI testing mIoU among all open-source LiDAR segmentation networks, we train the SalsaNext on the nuScenes dataset from scratch as there are no available pretrained nuScenes segmentation networks.

\subsection{Setup}

\textbf{Metrics:} We use mean intersection over union (mIoU) to evaluate the performance of semantic segmentation. For panoptic segmentation, we use the panoptic quality (PQ) metric~\cite{Kirillov_2019_CVPR}, defined as

\begin{equation}
    \vspace{-5pt}
    PQ = \underbrace{\frac{\sum_{\textbf{TP}}\textbf{IoU}}{|\textbf{TP}|}}_\text{SQ}\underbrace{\frac{|\textbf{TP}|}{|\textbf{TP}|+\frac{1}{2}|\textbf{TN}|+\frac{1}{2}|\textbf{FP}|}}_\text{RQ}.
\end{equation}

For an instance prediction to be considered as a TP, it needs at least $50\%$ overlap with the groundtruth. Recognition quality (RQ) shows the accuracy of finding TP, while Semantic quality (SQ) shows the average IoU in all TPs.  In addition, we report PQ$^{\dagger}$, which is proposed by Porzi \textit{et al.}~\cite{Porzi_2019_CVPR} to use only SQ as PQ in ``stuff'' classes. We also report the inference time to generate a single scan prediction. 

\textbf{Implementation details:} Following the same configuration of PolarNet~\cite{Zhang_2020_CVPR}, we discretize the 3D space within $[distance: 3 \sim 50\text{m}, z:-3 \sim 1.5\text{m}]$ to $[480,360,32]$ voxels in SemanticKITTI. We generate the groundtruth heatmap for the center prediction in a $\pm3*5$ window around the mass center of points, and correspondingly use the NMS with kernel size $\sigma=5$, threshold 0.1, and $k=100$ during the panoptic fusion. Compared to SemanticKITTI, nuScenes uses the LiDAR sensor that contains 32 beams rather than 64 beams. Furthermore, each scan in nuScenes has 34720 points and 34 instances on average, whereas SemanticKITTI has 104452 points and 5.3 instances. As a result, object points are more sparse in the nuScenes dataset. Hence, we consider an instance with a minimal of 20 points instead of 50 points as a valid instance during the panoptic segmentation evaluation. We use the same implementation setting as in SemanticKITTI to train Panoptic-PolarNet in nuScenes, except that we change the 3D space range to $[distance: 0 \sim 50\text{m}, z:-5 \sim 3\text{m}]$.

We implemented Panoptic-PolarNet in Pytorch on a single NVIDIA TITAN Xp GPU. We use the Adam optimizer with the default configuration. We use the combination of cross-entropy loss ($\mathcal{L}_{ce}$) and Lovasz softmax loss~\cite{berman2018lovasz} ($\mathcal{L}_{ls}$) to train our semantic segmentation head. For the instance head, we use the MSE loss ($\mathcal{L}_{hm}$) for the heatmap regression and L1 loss ($\mathcal{L}_{os}$) for the offset regression. The final loss is

\begin{equation}
    \mathcal{L} = \mathcal{L}_{ce}+\mathcal{L}_{ls} + \lambda_{hm}\mathcal{L}_{hm} + \lambda_{os}\mathcal{L}_{os},
\end{equation}
 where we set $\lambda_{hm} = 100$ and $\lambda_{os} = 10$. In addition to instance augmentation, during the training, we also use data augmentations, which randomly reflects a point cloud along $x$, $y$ and $x + y$ axis and randomly rotates the point cloud around the $Z$ axis. We apply dropblock~\cite{ghiasi2018dropblock} at the end of each up layer to further regularize the training of the proposed Panoptic-PolarNet. Unless specifically mentioned, all hyperparameters (percentage of points pruned per frame, \emph{etc}\onedot) are tuned on the validation dataset.
 
\begin{table*}
\centering
\caption{Panoptic segmentation results on the \textbf{validation} split of nuScenes.}
\label{tab:panoptic_val_nuscenes}
\resizebox{\linewidth}{!}{
\begin{tabular}{|l|c|cccc|ccc|ccc|c|}
\hline
Method                     & Latency  & PQ     & PQ$^{\dagger}$     & RQ     & SQ     & PQ$^{Th}$     & RQ$^{Th}$     & SQ$^{Th}$     & PQ$^{St}$     & RQ$^{St}$     & SQ$^{St}$  & mIoU\\ \hline
PolarNet~\cite{Zhang_2020_CVPR}  +  CBGS~\cite{zhu2019class}     & 0.208s   & 66.6\% & 70.3\% & 78.0\% & 84.6\% & 63.8\% & 74.1\% & 85.1\% & 71.4\% & 84.5\% & 83.7\% & 71.8\%\\
SalsaNext~\cite{cortinhal2020salsanext} + CBGS~\cite{zhu2019class}      & 0.207s   & 61.6\% & 66.3\% & 72.3\% & 84.5\% & 59.5\% & 68.1\% & 86.6\% & 65.1\% & 79.5\% & 81.0\% & 63.4\%\\
Panoptic-PolarNet        & 0.099s   & 67.7\% & 71.0\% & 78.1\% & 86.0\% & 65.2\% & 74.0\% & 87.2\% & 71.9\% & 84.9\% & 83.9\% & 69.3\%\\ \hline
\end{tabular}
}
\end{table*}

\subsection{Quantitative Results}

Table~\ref{tab:panoptic_test_semantickitti} shows the comparison between Panoptic-PolarNet and the baselines on the test split of SemanticKITTI. Our method outperforms the best baseline by 1.4\% in PQ while having a near real-time inference speed, setting a new state-of-the-art performance for the LiDAR panoptic segmentation. It is noticeable that our method has a significant improvement for the ``thing'' classes compared with other dedicated state-of-the-art LiDAR object detector. We credit our superior instance prediction to the architecture design and augmentation methods. On the other hand, the results for ``stuff'' classes show a very close correlation to the semantic segmentation. Nevertheless, we still manage to achieve a better performance than the best combining baseline method because of a more well-balanced segmentation results among all classes. Our mini version of Panoptic-PolarNet with $[320,240,32]$ grid size achieves a comparable result and only need 2/3 of inference time. More detailed results with respect to each class will be presented in the supplemental materials.

We present our SemanticKITTI validation results in Table~\ref{tab:panoptic_val_semantickitti}. We additionally experimented with different settings of Panoptic-PolarNet with more variants of combining baselines. Similar to \cite{Zhang_2020_CVPR}, we found that polar coordinate prevails Cartesian coordinate in terms of every metric while having a slower inference time. All three settings of Panoptic-PolarNet outperform the best baseline method by a large margin.

We report the result on the validation set of nuScenes in Table~\ref{tab:panoptic_val_nuscenes}. Our method outperforms the combining baseline method by 1.1\% in PQ with only half of the time. However, due to the increase in the number of instances, the inference time in nuScenes is slightly higher than SemanticKITTI. 

\begin{table}
\centering
\caption{Ablation study of Panoptic PolarNet on the \textbf{validation} split of SemanticKITTI. `SU',`IO',`IGA',`ILA',`SAP', `Vis' stand for training with first three up layers shared among semantic and instance heads (SU), instance oversampling (IO), instance global augmentation (IGA), instance local augmentation (ILA), self adversarial pruning (SAP), and visibility feature (Vis).}
\vspace{-10pt}
\label{tab:ablation_study}
\resizebox{\linewidth}{!}{
\begin{tabular}{ccccccc|cc}
\hline
SU  & IO & IGA & $\sigma$ = 5 & ILA & SAP & Vis & PQ & mIoU   \\

\hline
   &    &    &    &    &   &   & 51.6\% & 57.8\%    \\
$\times$  &    &    &    &    &  &   & 52.3\% & 58.1\%    \\
$\times$  & $\times$  &    &    &    &  &   & 55.1\% & 60.2\%     \\
$\times$ & $\times$  & $\times$  &    &    &  &   & 57.0\% & 62.1\%   \\
$\times$  &$\times$  & $\times$  & $\times$  &    &  &   & 57.3\% & 61.3\%    \\
$\times$ & $\times$  &   & $\times$  & $\times$  &   &  & 57.3\% & 61.7\% \\
$\times$ & $\times$  & $\times$  & $\times$  & $\times$  &   &  & 57.4\% & 61.6\% \\
$\times$ & $\times$  & $\times$  & $\times$  & $\times$  & $\times$ &   & 57.5\% & 62.1\% \\
$\times$ & $\times$  & $\times$  & $\times$  & $\times$  &  & $\times$  & 58.0\% & 62.6\% \\
$\times$ & $\times$  & $\times$  & $\times$  & $\times$  & $\times$ & $\times$  & 59.1\% & 64.5\% \\
\hline
\end{tabular}
}
\vspace{-5pt}
\end{table}

\begin{table}
\centering
\caption{Oracle test of Panoptic-PolarNet on the \textbf{validation} split of SemanticKITTI.}
\vspace{-10pt}
\label{tab:oracle test}
\resizebox{\linewidth}{!}{
\begin{tabular}{ccc|cc}
\hline
GT Heatmap  & GT Offset & GT Semantic & PQ & mIoU          \\

\hline
         &          &          & 59.1\% & 64.5\%    \\
$\times$ &          &          & 59.5\% & 64.5\%    \\
         & $\times$ &          & 59.4\% & 64.5\%    \\
$\times$ & $\times$ &          & 60.1\% & 64.5\%    \\
         &          & $\times$ & 91.9\% & 94.1\%    \\
$\times$ & $\times$ & $\times$ & 96.8\% & 96.4\%    \\

\hline
\end{tabular}
}
\vspace{-5pt}
\end{table}

\subsection{Ablation Studies}
To further analyze the influence of each component, we conducted the ablation studies on the validation split of SemanticKITTI, as shown in Table~\ref{tab:ablation_study}. We started by training Panoptic-PolarNet without any augmentation and used two independent decoding networks for semantic and instance heads. Rather than using an attention module to connect the learned information between semantic and instance heads, we found out that directly sharing the first three decoding layers can increase the PQ from 51.6\% to 52.3\% with an even smaller computation load. This indicates that the features learned by semantic and instance heads share plenty of similarities in our setting. Next, we tested the effect of different instance augmentation components on the segmentation results. Instance oversampling improves PQ by 2.8\% and mIoU by 2.1\%, which benefits the ``thing'' classes that seldom appear in a scan most. On the other hand, instance global augmentation and local augmentation both have improvements, and using all three instance augmentation methods gives the best result in PQ. Self-adversarial pruning slightly improves the results in terms of PQ but helps to stabilize the semantic results, especially for ``stuff'' classes. Lastly, visibility feature improves the PQ by 1.6\%. Those classes that are mostly surrounded by visible space, like bicyclist and motorcyclist, benefit most from the visibility feature.

We also conducted an oracle test, as shown in Table~\ref{tab:oracle test} to investigate the room for improvement in Panoptic-PolarNet. We replaced some predictions in the semantic and instance heads to the ground truth for each experiment and generated the panoptic predictions using the same fusion step. It can be seen that our heatmap and offset prediction are both very close to the ground truth in our test setting and, when combined, have only 1.0\% difference in PQ compared to the ground truth instance clustering. Conversely, ground truth semantic prediction greatly impacts the results and increases both the PQ and mIoU to above 90\%. This matches the finding in ~\cite{cheng2020panoptic} that the biggest bottleneck in proposal-free panoptic segmentation is the semantic segmentation. Lastly, Table~\ref{tab:oracle test} shows that PQ and mIoU are 96.8\% and 96.4\% when we use all three ground truth together. This shows that the discretization and projection errors are relatively small in our setting and also verifies our assumption that it is sufficient to separate instances directly on the BEV. 

\subsection{Runtime}
We report the detailed runtime and model size of Panoptic-PolarNet with different settings in Table~\ref{tab:runtime}. Compared with PolarNet~\cite{Zhang_2020_CVPR} that solves only for semantic segmentation, our method merely increases the parameter size by 0.1M, and prediction time by 0.02s. Such an insignificant increase reflects our method's high efficiency in generating instance prediction on top of a well-established semantic segmentation network. The inference time difference mostly comes from the fusion step, while it is worth noting that this part has a big room for improvement if better optimized. Both Panoptic-PolarNet and its mini version can process LiDAR data in real-time as a typical LiDAR sensor works at 10 FPS~\cite{geiger2012cvpr,sun2020scalability}.

\begin{table}
\centering
\caption{Runtime and parameter size comparisons of Panoptic-PolarNet.}
\vspace{-8pt}
\label{tab:runtime}
\resizebox{\linewidth}{!}{
\begin{tabular}{l|cccc}
\hline
Model  & Pred & Fusion & Params & FPS          \\

\hline
PolarNet        &   0.059s      &  -    & 13.6M  & 16.9    \\
Panoptic-PolarNet &  0.061s  &  0.025s    & 13.7M &  11.6   \\
Panoptic-PolarNet-mini &  0.043s  & 0.014s   & 13.7M & 17.5  \\

\hline
\end{tabular}
}
\vspace{-5pt}
\end{table}

%% file: conclusion.tex
\section{Conclusion}

In this paper, we present a real-time proposal-free LiDAR point cloud panoptic segmentation framework named Panoptic-PolarNet. Our method builds upon the established semantic segmentation network and solves the instance segmentation by center regression on the polar BEV map. This design highly simplifies the complexity of the panoptic segmentation, requiring a negligible computation overhead on top of the semantic segmentation and achieving the state-of-the-art result on both SemanticKITTI and nuScenes datasets. We also propose several novel augmentation methods that can be generalized to any other LiDAR point cloud segmentation methods. We hope Panoptic-PolarNet can serve as a strong baseline for future research and a helpful framework for current semantic segmentation methods to migrate to the panoptic segmentation.

%% file: supp_arxiv.tex
\section{Supplementary Material}

\subsection{Discussion}

\begin{table*}[!htbp]
	\centering
	\caption{Class-wise results on \textbf{test} split of SemanticKITTI.}
	\label{tab:SemanticKITTI_class}
	\resizebox{\linewidth}{!}{%
	\begin{tabular}{|c|*{19}{c}|c|}
		\hline
		 metrics &\rotatebox{90}{car} &	\rotatebox{90}{bicycle} &	\rotatebox{90}{motorcycle} &	\rotatebox{90}{truck}&	\rotatebox{90}{other-vehicle} &	\rotatebox{90}{person} &	\rotatebox{90}{bicyclist} &	\rotatebox{90}{motorcyclist} &	\rotatebox{90}{road} &	\rotatebox{90}{parking} & \rotatebox{90}{sidewalk} &	\rotatebox{90}{other-ground} &	\rotatebox{90}{building} &	\rotatebox{90}{fence} &	\rotatebox{90}{vegetation} &	\rotatebox{90}{trunk} &	\rotatebox{90}{terrain} &	\rotatebox{90}{pole} &	\rotatebox{90}{traffic-sign}&\rotatebox{90}{mean} \\
		\hline
		
		PQ & 88.8\% & 33.0\% &  51.8\% & 35.2\% & 37.6\% & 57.3\% & 69.9\% & 52.4\% &  88.6\% & 42.6\% & 61.2\% &  1.6\% &  85.7\% & 46.0\% & 75.7\% & 54.5\% & 41.5\% &  48.5\% &  56.4\% & 54.1\%\\
		RQ & 96.2\% & 46.7\% &  59.9\% & 38.7\% & 41.5\% & 65.7\% & 78.6\% & 57.7\% &  96.8\% & 56.2\% & 76.3\% &  2.8\% &  92.1\% & 62.3\% & 92.5\% & 73.6\% & 55.5\% &  66.0\% &  75.2\% & 65.0\%\\
		SQ & 92.3\% &  70.7\% &  86.4\% & 90.9\% & 90.5\% & 87.2\% & 88.9\% & 90.9\% &  91.5\% & 75.9\% & 80.3\% &  55.6\% &  93.0\% & 73.8\% & 81.9\% & 74.1\% & 74.9\% & 73.4\% &  75.0\% & 81.4\%\\
		IoU & 94.4\% & 38.7\% &  48.2\% & 46.2\% & 34.5\% & 51.1\% & 63.9\% & 24.9\% &  90.8\% & 61.3\% & 74.6\% &  16.5\% &  89.9\% & 61.1\% & 83.4\% & 66.7\% & 68.0\% & 56.8\% &  58.5\% & 59.5\%\\
		
		\hline
	\end{tabular}
	}
\vspace{0pt}
\end{table*}

\begin{table*}[!htbp]
	\centering
	\caption{Class-wise results on \textbf{validation} split of nuScenes.}
	\label{tab:nuScenes_class}
	\resizebox{\linewidth}{!}{%
	\begin{tabular}{|c|*{16}{c}|c|}
		\hline
		 metrics &\rotatebox{90}{barrier} &	\rotatebox{90}{bicycle} &	\rotatebox{90}{bus} &	\rotatebox{90}{car}&	\rotatebox{90}{construction vehicle} &	\rotatebox{90}{motorcycle} &	\rotatebox{90}{pedestrian} &	\rotatebox{90}{traffic cone} &	\rotatebox{90}{trailer} &	\rotatebox{90}{truck} & \rotatebox{90}{driveable surface} &	\rotatebox{90}{other flat} &	\rotatebox{90}{sidewalk} &	\rotatebox{90}{terrain} &	\rotatebox{90}{manmade} &	\rotatebox{90}{vegetation} &\rotatebox{90}{mean} \\
		\hline
		
		PQ & 41.5\% & 58.0\% &  70.4\% & 89.0\% & 36.4\% & 78.4\% & 85.1\% & 80.7\% &  49.7\% & 63.3\% & 95.7\% &  53.9\% &  67.7\% & 49.8\% & 84.3\% & 80.0\% & 67.7\%\\
		RQ & 54.4\% & 68.6\% &  76.1\% & 95.9\% & 44.9\% & 86.0\% & 95.2\% & 91.8\% &  58.2\% & 69.4\% & 100.0\% &  66.3\% &  85.3\% & 65.1\% & 98.2\% & 94.7\% & 78.1\%\\
		SQ & 76.3\% &  84.6\% & 92.5\% & 92.8\% & 81.2\% & 91.2\% & 89.4\% & 87.9\% &  85.3\% & 91.3\% & 95.7\% &  81.3\% &  79.4\% & 76.5\% & 86.0\% & 84.5\% & 86.0\%\\
		IoU & 52.3\% & 28.1\% & 88.0\% & 90.3\% & 32.4\% & 71.7\% & 72.2\% & 52.8\% &  58.3\% & 76.6\% & 95.9\% &  68.8\% &  74.3\% & 73.4\% & 87.3\% & 85.5\% & 69.3\%\\
		
		\hline
	\end{tabular}
	}
\vspace{0pt}
\end{table*}

\textbf{Choice of the loss:} We adopted a combination of cross-entropy loss and Lovasz softmax loss in the semantic head. Given the highly imbalanced class distribution in LiDAR point clouds, the cross-entropy loss will favor those classes that are the majority of points, like the road class and the building class. Conversely, Lovasz softmax loss optimizes directly on the mIoU Jaccard index, which treats all classes equally. Combining these two losses will force the network to optimize toward an overall accurate prediction while focusing more on hard classes. In the instance head, we chose the MSE loss instead of focal loss~\cite{lin2017focal} for the heatmap regression. The reason is that we do not necessarily need a very accurate prediction of the center in the BEV due to the scarcity of instance overlaps. However, we need a monotonically decreasing heatmap from the center to the edge to have a proper keypoint selection in the NMS. Also, the experiments showed that focal loss decreases the PQ by 1.3\%.

\textbf{Data augmentation:} We apply instance oversampling to compensate for two imbalances in the LiDAR point cloud: (1) The imbalance between ``thing'' and ``stuff''. Points belong to ``thing'' classes usually consist of only a small portion of the point cloud. (2) The imbalance between different ``thing'' classes. For example, the most occurring class, car, has around $10^{7}$ time more points than the least occurring class, motorcyclist in the SemanticKITTI dataset. Experiments show that even though this oversampling will decrease the segmentation accuracy in ``stuff'', the overwhelmingly increases in ``thing'' can still provide a huge improvement to the PQ and mIoU. Our experiments also show that either simply putting instance points at any place in a point cloud or transform it through its center will decrease the PQ. We conclude that such simple augmentation ignores projection properties, introduces inconsistency into the LiDAR point cloud, and thus entangles BEV feature learning.

\textbf{Proposal-free vs. proposal-based:} Even though proposal-based panoptic segmentation methods dominate in the 2D domain, there are only a few existing approaches for LiDAR point clouds. We think there are two reasons. First, proposal-based methods rely heavily on the annotation of bounding boxes, whereas point cloud datasets do not necessarily provide such annotations. Second, most current proposal-based object detection methods, like what we assume in our instance head, are not designed to represent the scene along the Z-axis. Lacking proper representation makes it more challenging to achieve a competitive result while maintaining speed when modified into a panoptic segmentation network. 

\textbf{End-to-end training:} We only train the network to get an intermediate result and use a majority voting fusion to generate the final panoptic segmentation. Making the proposal-free panoptic segmentation network end-to-end trainable is still an open problem to explore in the future. 

\begin{figure}
\centering
    \begin{subfigure}[b]{0.49\linewidth}
        \centering
        \includegraphics[width=\textwidth]{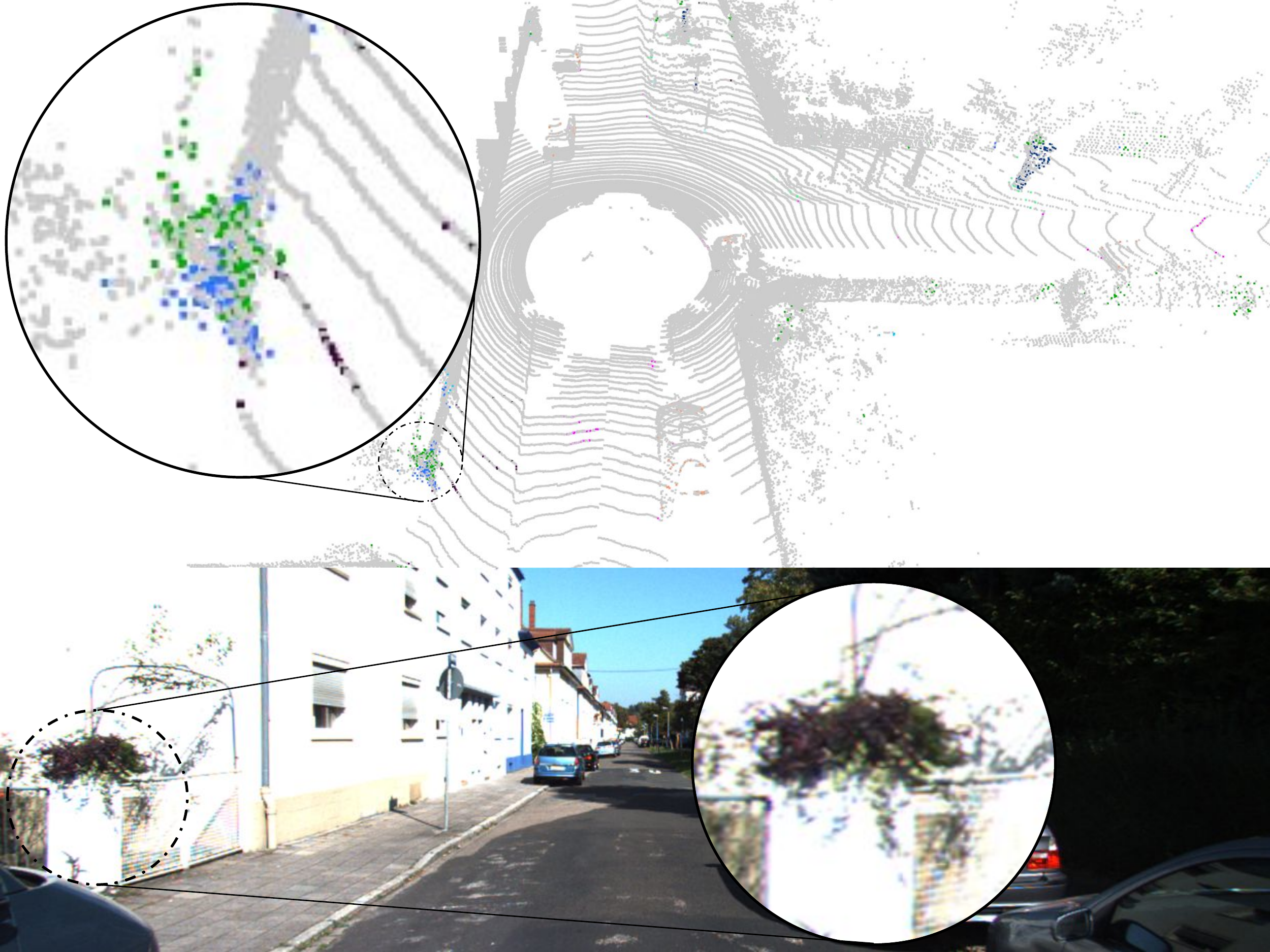}
    \end{subfigure}
    \begin{subfigure}[b]{0.49\linewidth}
        \centering
        \includegraphics[width=\textwidth]{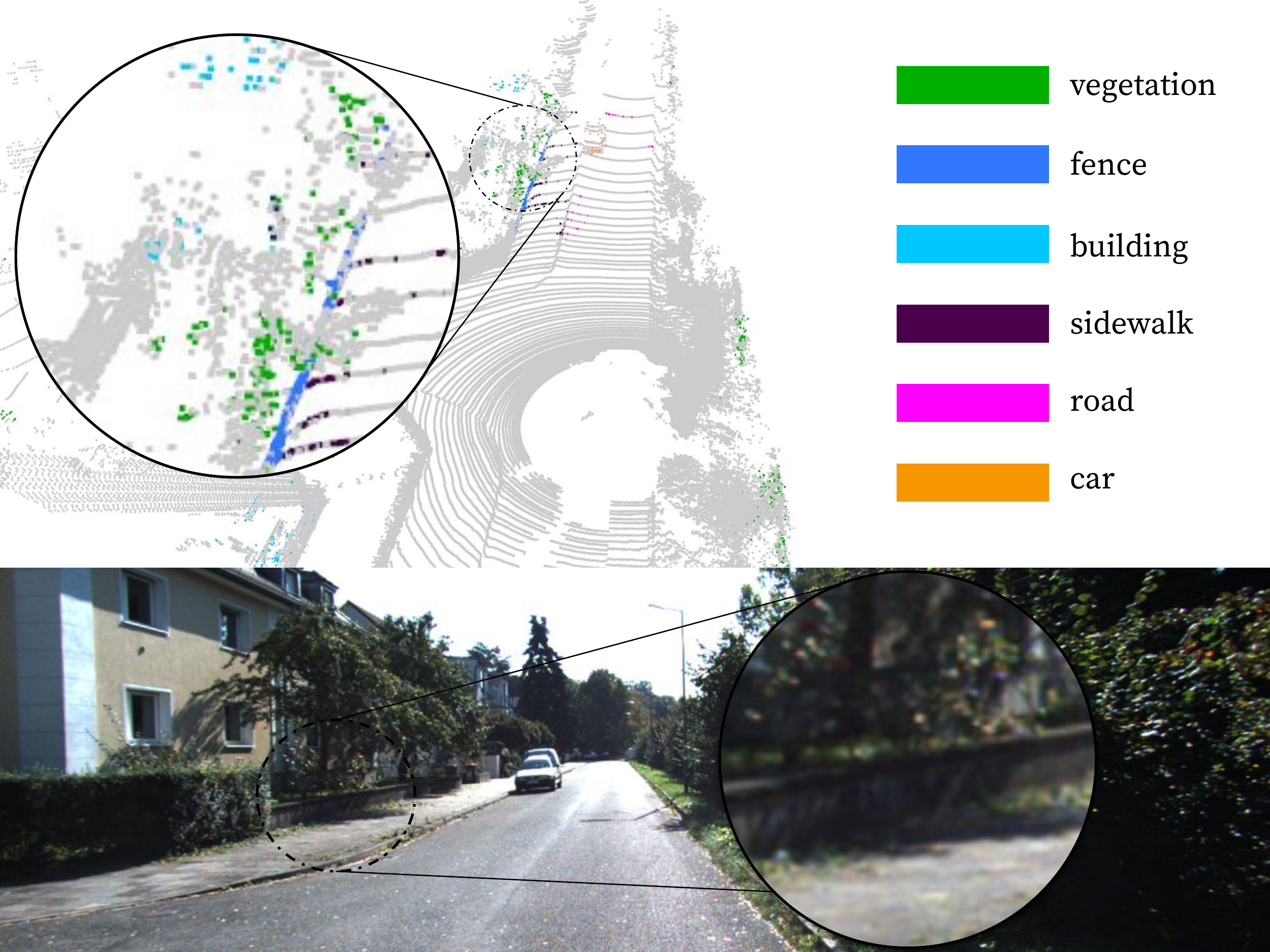}
    \end{subfigure}
\caption{We highlight SAP-pruned points in color. Left: SAP prunes tangled vegetation and fence; Right: SAP prunes garden curbs that are annotated as the `` fence ''}
\label{fig:pruning_vis}
\end{figure}

\textbf{Self-adversarial Pruning Visualization:} SAP is designed to remove ambiguous, noisy or/ and informative points. Since SemanticKITTI is a well-annotated dataset, we visually find SAP tends to remove challenging or ambiguous cases. Some examples are shown in Fig.~\ref{fig:pruning_vis}.

\subsection{Class-wise Results}
We show the class-wise results of Panoptic-PolarNet on SemanticKITTI and nuScenes in Table~\ref{tab:SemanticKITTI_class} and Table~\ref{tab:nuScenes_class}. Our method has a similar panoptic segmentation performance in the corresponding classes among these two datasets. The low performance comes from the class that either has a small physical shape (like bicycle) or has a small number of instances in the dataset (like truck and construction vehicle). Despite being a more challenging dataset due to its significantly higher number of instances, nuScenes has fewer classes than SemanticKITTI, which makes it more distinguishable and thus having higher PQ and mIoU.

\subsection{Qualitative Results}
\begin{figure*}
\centering
    \begin{subfigure}[t]{0.24\textwidth}
        \centering
        \includegraphics[width=\textwidth]{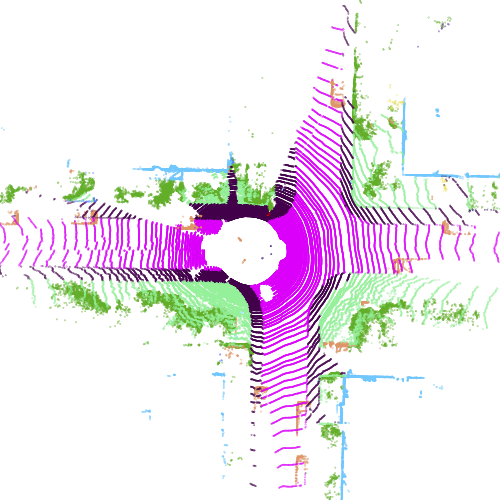}
    \end{subfigure}
    \hspace*{\fill}%
    \begin{subfigure}[t]{0.24\textwidth}
        \centering
        \includegraphics[width=\textwidth]{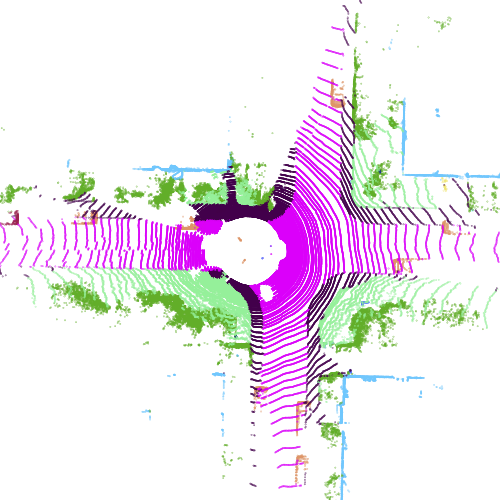}
    \end{subfigure}
    \hspace*{\fill}%
    \begin{subfigure}[t]{0.24\textwidth}
        \centering
        \includegraphics[width=\textwidth]{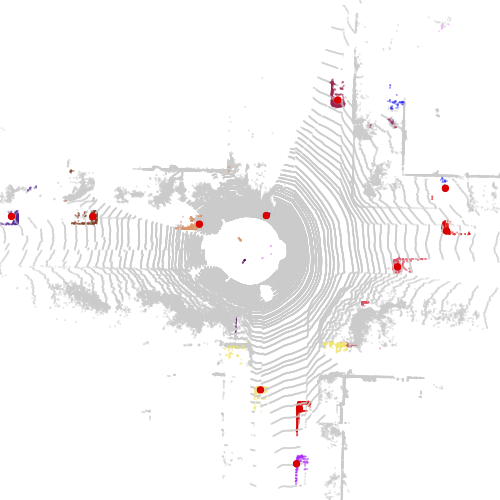}
    \end{subfigure}
    \hspace*{\fill}%
    \begin{subfigure}[t]{0.24\textwidth}
        \centering
        \includegraphics[width=\textwidth]{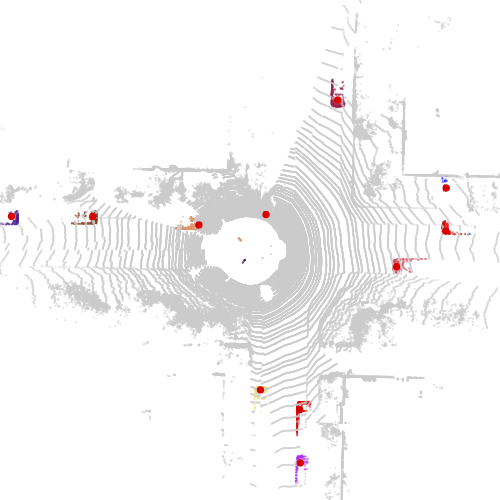}
    \end{subfigure}
    
    \begin{subfigure}[t]{0.24\textwidth}
        \centering
        \includegraphics[width=\textwidth]{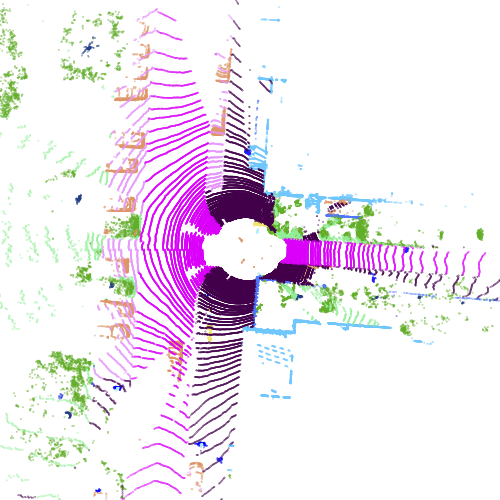}
    \end{subfigure}
    \hspace*{\fill}%
    \begin{subfigure}[t]{0.24\textwidth}
        \centering
        \includegraphics[width=\textwidth]{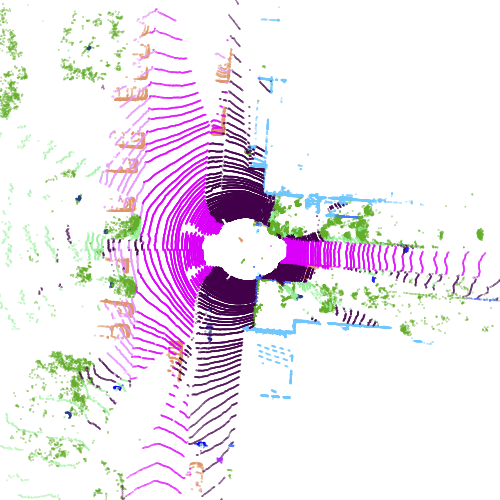}
    \end{subfigure}
    \hspace*{\fill}%
    \begin{subfigure}[t]{0.24\textwidth}
        \centering
        \includegraphics[width=\textwidth]{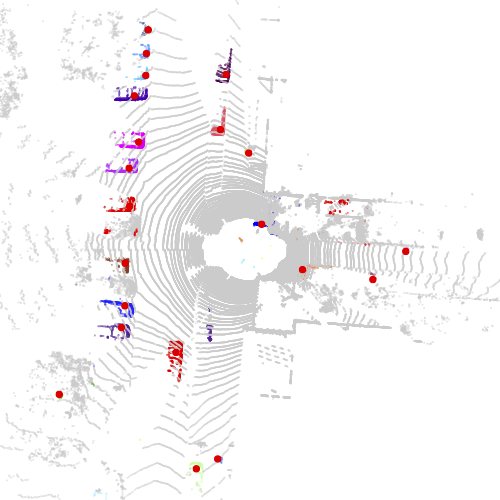}
    \end{subfigure}
    \hspace*{\fill}%
    \begin{subfigure}[t]{0.24\textwidth}
        \centering
        \includegraphics[width=\textwidth]{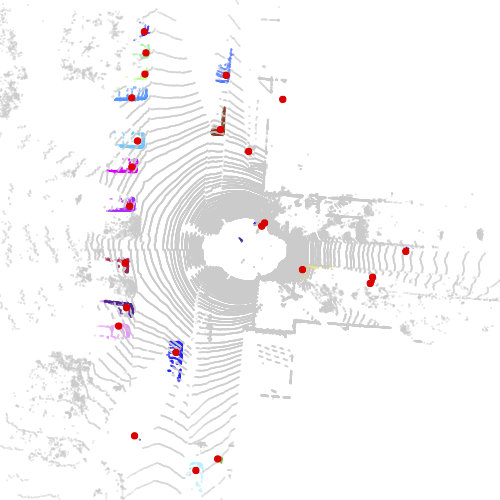}
    \end{subfigure}
    
    \begin{subfigure}[t]{0.24\textwidth}
        \centering
        \includegraphics[width=\textwidth]{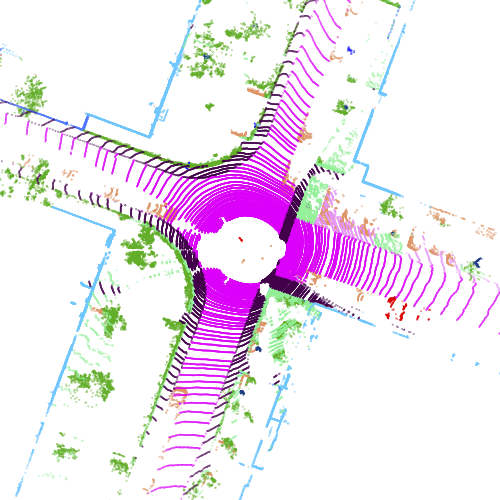}
    \end{subfigure}
    \hspace*{\fill}%
    \begin{subfigure}[t]{0.24\textwidth}
        \centering
        \includegraphics[width=\textwidth]{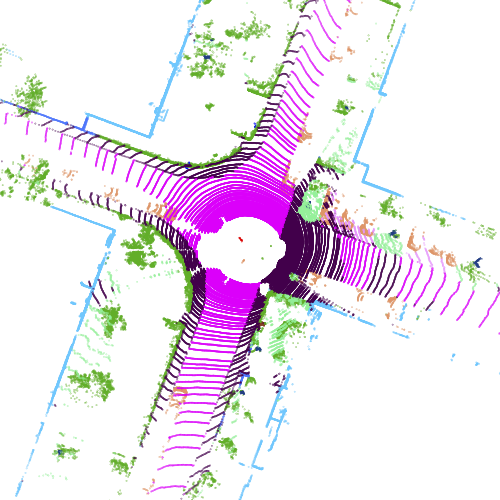}
    \end{subfigure}
    \hspace*{\fill}%
    \begin{subfigure}[t]{0.24\textwidth}
        \centering
        \includegraphics[width=\textwidth]{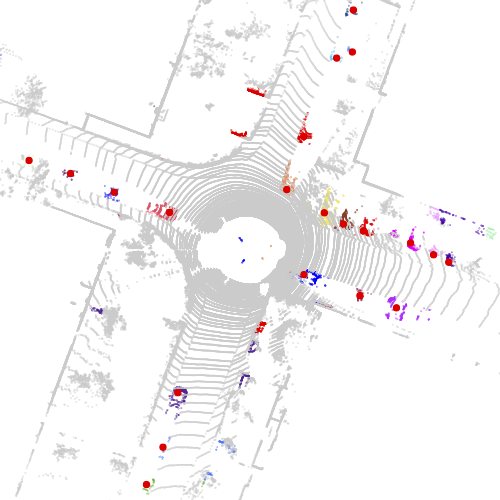}
    \end{subfigure}
    \hspace*{\fill}%
    \begin{subfigure}[t]{0.24\textwidth}
        \centering
        \includegraphics[width=\textwidth]{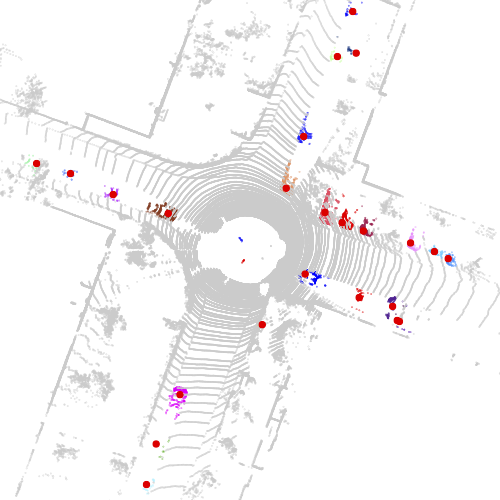}
    \end{subfigure}
    
    \begin{subfigure}[t]{0.24\textwidth}
        \centering
        \includegraphics[width=\textwidth]{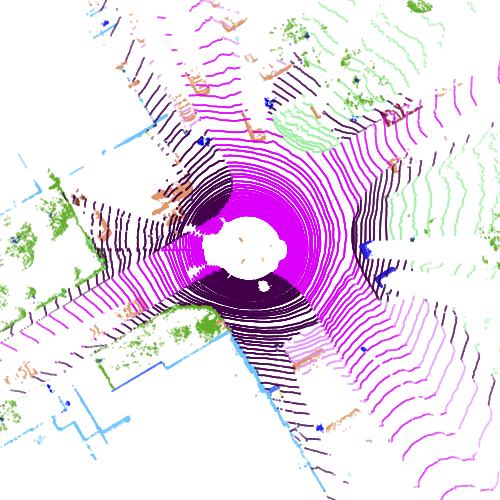}
    \end{subfigure}
    \hspace*{\fill}%
    \begin{subfigure}[t]{0.24\textwidth}
        \centering
        \includegraphics[width=\textwidth]{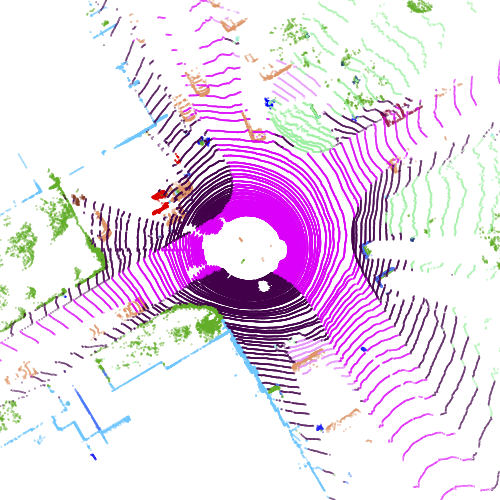}
    \end{subfigure}
    \hspace*{\fill}%
    \begin{subfigure}[t]{0.24\textwidth}
        \centering
        \includegraphics[width=\textwidth]{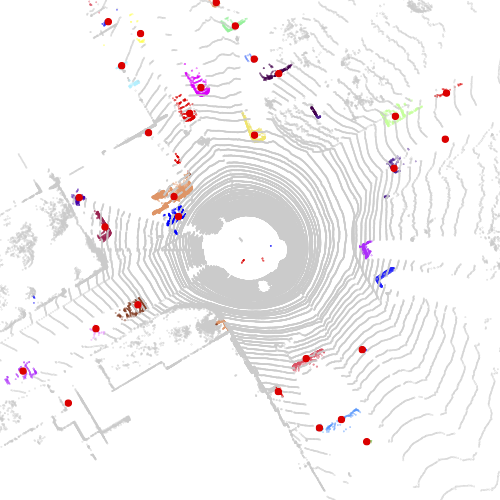}
    \end{subfigure}
    \hspace*{\fill}%
    \begin{subfigure}[t]{0.24\textwidth}
        \centering
        \includegraphics[width=\textwidth]{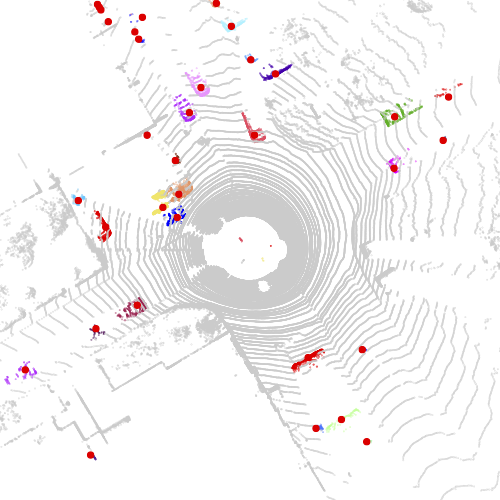}
    \end{subfigure}
    
    \begin{subfigure}[t]{0.24\textwidth}
        \centering
        \includegraphics[width=\textwidth]{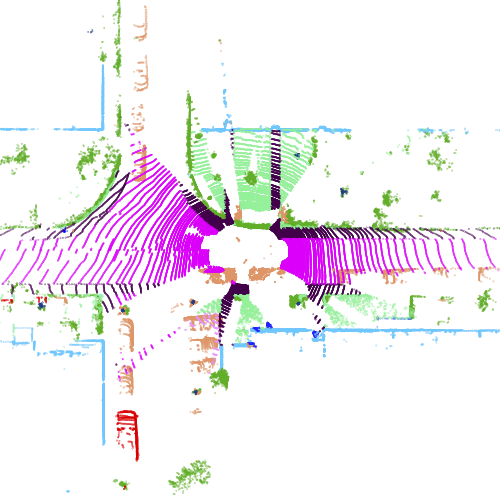}
        \caption{Semantic Ground Truth}
    \end{subfigure}
    \hspace*{\fill}%
    \begin{subfigure}[t]{0.24\textwidth}
        \centering
        \includegraphics[width=\textwidth]{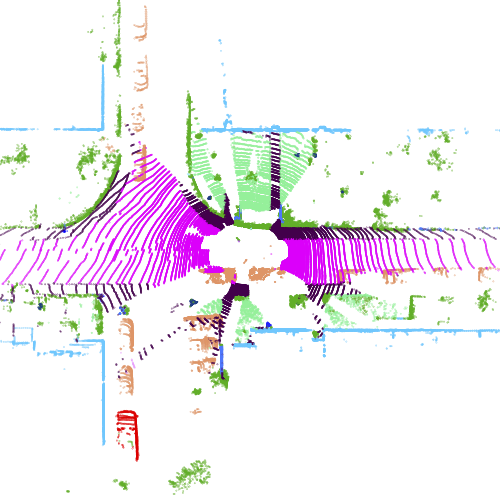}
        \caption{Semantic Prediction}
    \end{subfigure}
    \hspace*{\fill}%
    \begin{subfigure}[t]{0.24\textwidth}
        \centering
        \includegraphics[width=\textwidth]{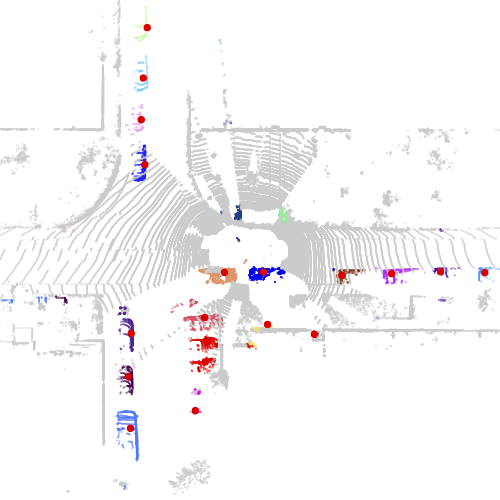}
        \caption{Instance Ground Truth}
    \end{subfigure}
    \hspace*{\fill}%
    \begin{subfigure}[t]{0.24\textwidth}
        \centering
        \includegraphics[width=\textwidth]{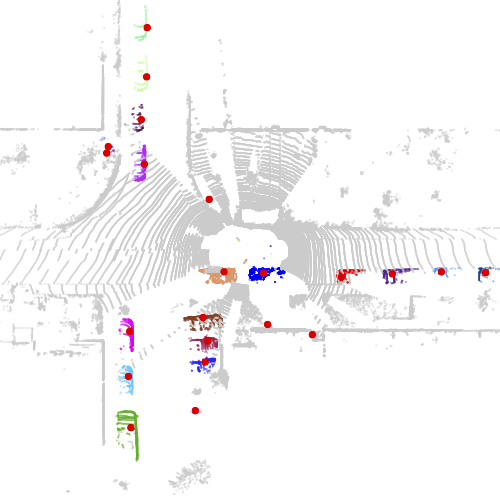}
        \caption{Instance Prediction}
    \end{subfigure}
    
\caption{Visualization of Panoptic-PolarNet on the SemanticKITTI dataset. The red dots in the instance prediction represent the center for each instance.}
\label{fig:semkitti_visualization}
\end{figure*}

\begin{figure*}
\centering
    \begin{subfigure}[t]{0.24\textwidth}
        \centering
        \includegraphics[width=\textwidth]{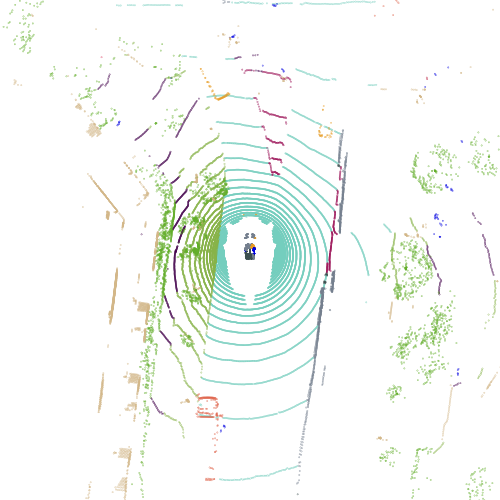}
    \end{subfigure}
    \hspace*{\fill}%
    \begin{subfigure}[t]{0.24\textwidth}
        \centering
        \includegraphics[width=\textwidth]{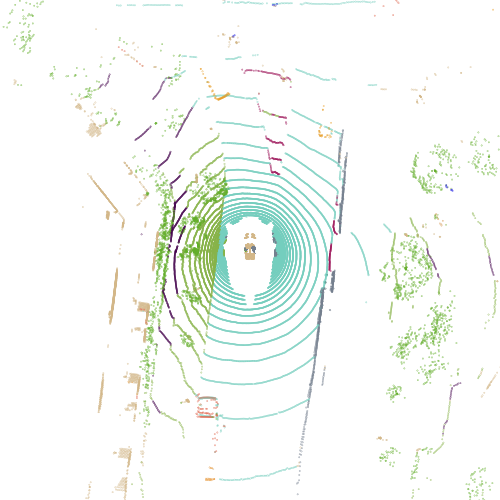}
    \end{subfigure}
    \hspace*{\fill}%
    \begin{subfigure}[t]{0.24\textwidth}
        \centering
        \includegraphics[width=\textwidth]{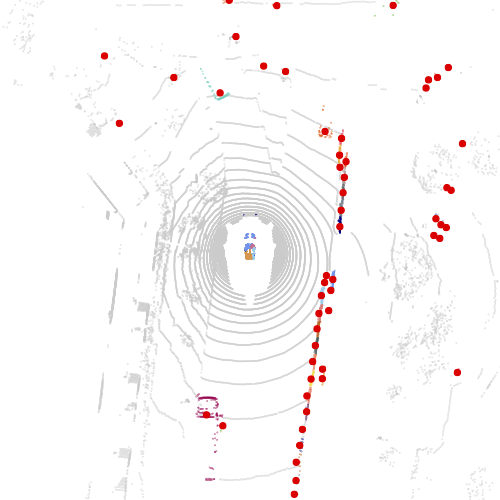}
    \end{subfigure}
    \hspace*{\fill}%
    \begin{subfigure}[t]{0.24\textwidth}
        \centering
        \includegraphics[width=\textwidth]{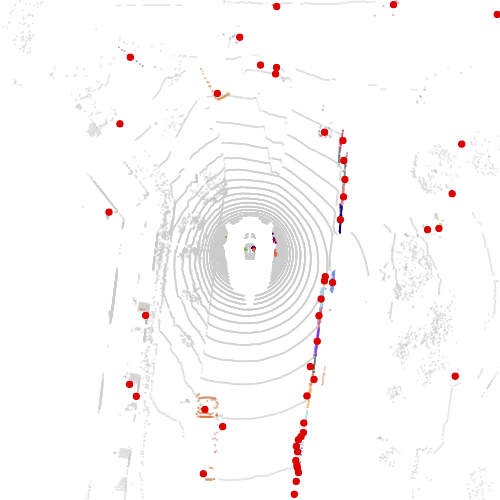}
    \end{subfigure}
    
    \begin{subfigure}[t]{0.24\textwidth}
        \centering
        \includegraphics[width=\textwidth]{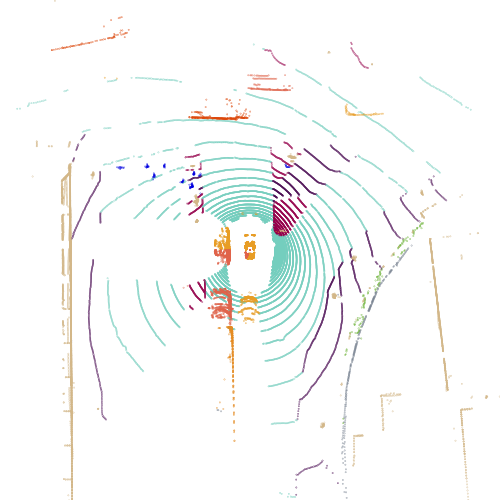}
    \end{subfigure}
    \hspace*{\fill}%
    \begin{subfigure}[t]{0.24\textwidth}
        \centering
        \includegraphics[width=\textwidth]{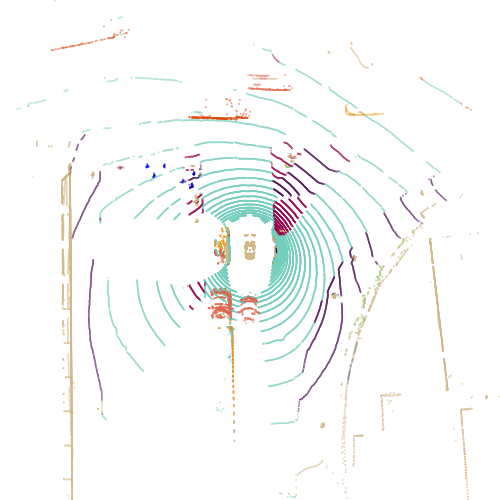}
    \end{subfigure}
    \hspace*{\fill}%
    \begin{subfigure}[t]{0.24\textwidth}
        \centering
        \includegraphics[width=\textwidth]{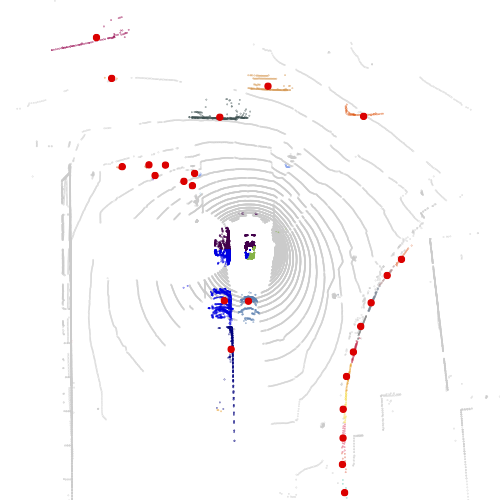}
    \end{subfigure}
    \hspace*{\fill}%
    \begin{subfigure}[t]{0.24\textwidth}
        \centering
        \includegraphics[width=\textwidth]{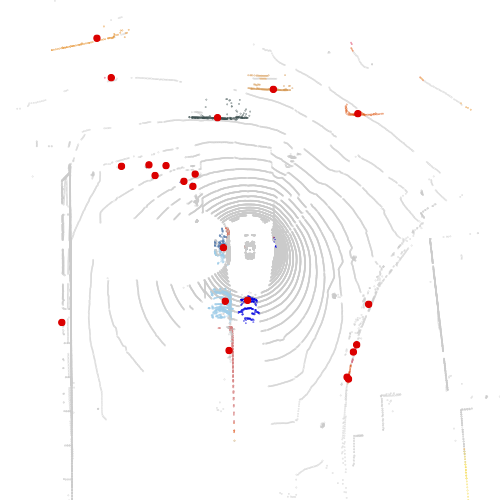}
    \end{subfigure}
    
    \begin{subfigure}[t]{0.24\textwidth}
        \centering
        \includegraphics[width=\textwidth]{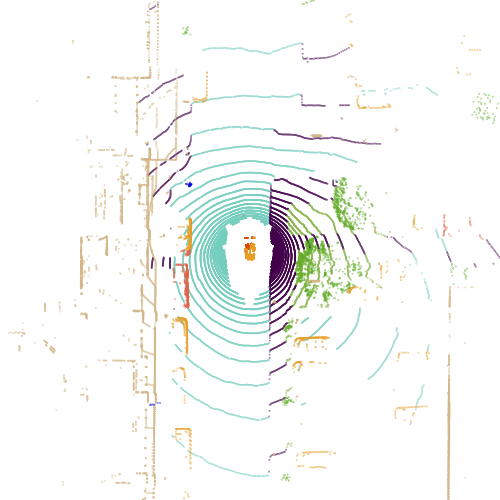}
    \end{subfigure}
    \hspace*{\fill}%
    \begin{subfigure}[t]{0.24\textwidth}
        \centering
        \includegraphics[width=\textwidth]{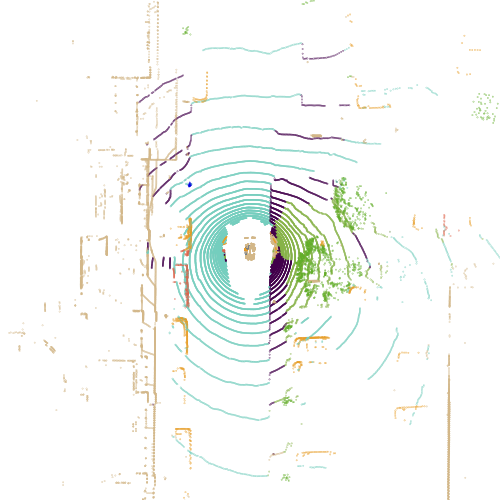}
    \end{subfigure}
    \hspace*{\fill}%
    \begin{subfigure}[t]{0.24\textwidth}
        \centering
        \includegraphics[width=\textwidth]{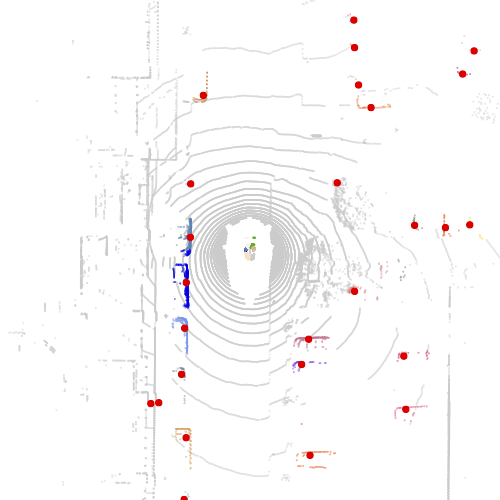}
    \end{subfigure}
    \hspace*{\fill}%
    \begin{subfigure}[t]{0.24\textwidth}
        \centering
        \includegraphics[width=\textwidth]{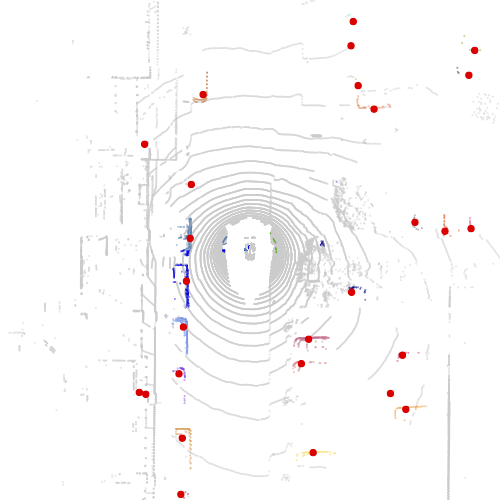}
    \end{subfigure}
    
    \begin{subfigure}[t]{0.24\textwidth}
        \centering
        \includegraphics[width=\textwidth]{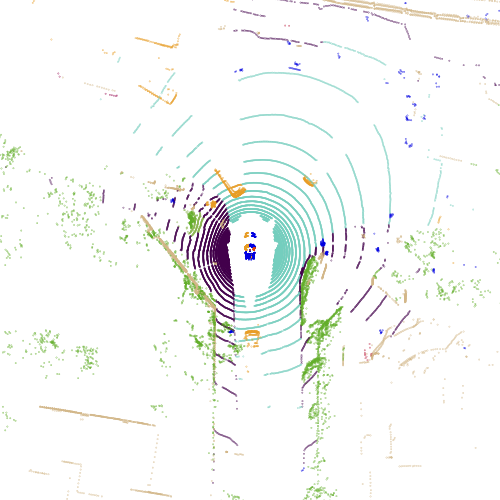}
    \end{subfigure}
    \hspace*{\fill}%
    \begin{subfigure}[t]{0.24\textwidth}
        \centering
        \includegraphics[width=\textwidth]{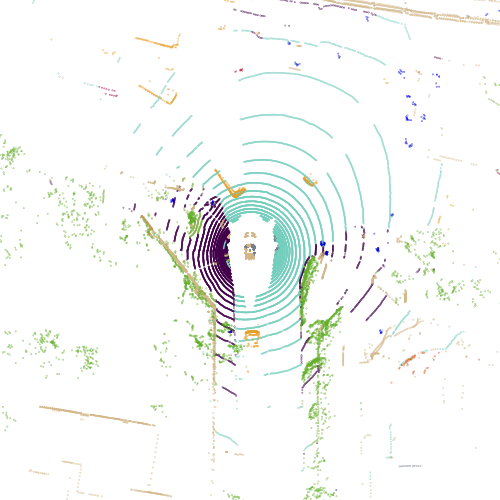}
    \end{subfigure}
    \hspace*{\fill}%
    \begin{subfigure}[t]{0.24\textwidth}
        \centering
        \includegraphics[width=\textwidth]{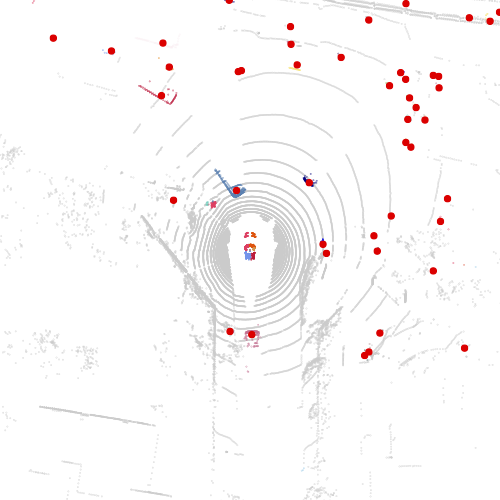}
    \end{subfigure}
    \hspace*{\fill}%
    \begin{subfigure}[t]{0.24\textwidth}
        \centering
        \includegraphics[width=\textwidth]{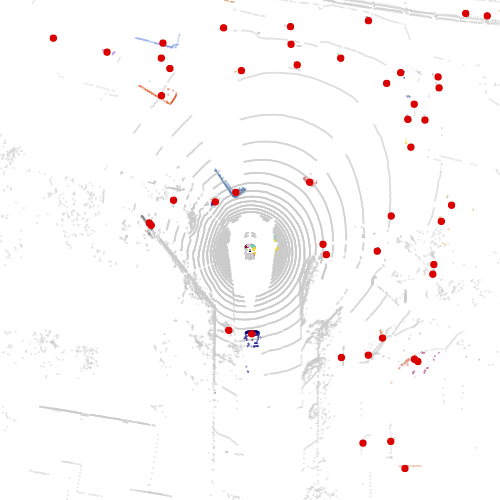}
    \end{subfigure}
    
    \begin{subfigure}[t]{0.24\textwidth}
        \centering
        \includegraphics[width=\textwidth]{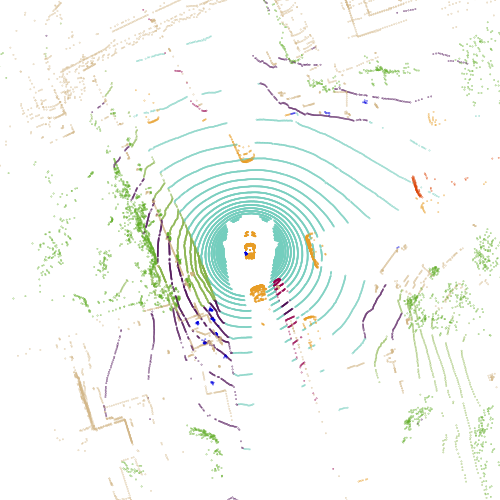}
        \caption{Semantic Ground Truth}
    \end{subfigure}
    \hspace*{\fill}%
    \begin{subfigure}[t]{0.24\textwidth}
        \centering
        \includegraphics[width=\textwidth]{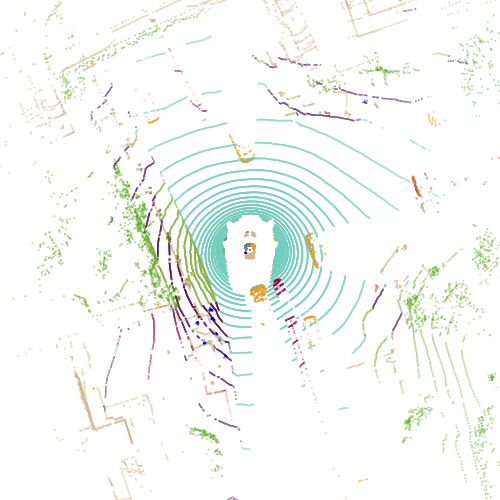}
        \caption{Semantic Prediction}
    \end{subfigure}
    \hspace*{\fill}%
    \begin{subfigure}[t]{0.24\textwidth}
        \centering
        \includegraphics[width=\textwidth]{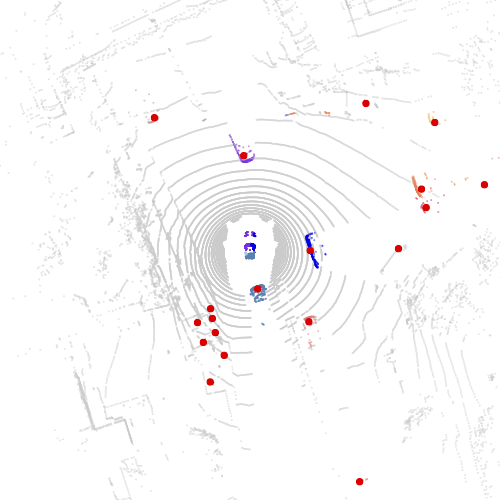}
        \caption{Instance Ground Truth}
    \end{subfigure}
    \hspace*{\fill}%
    \begin{subfigure}[t]{0.24\textwidth}
        \centering
        \includegraphics[width=\textwidth]{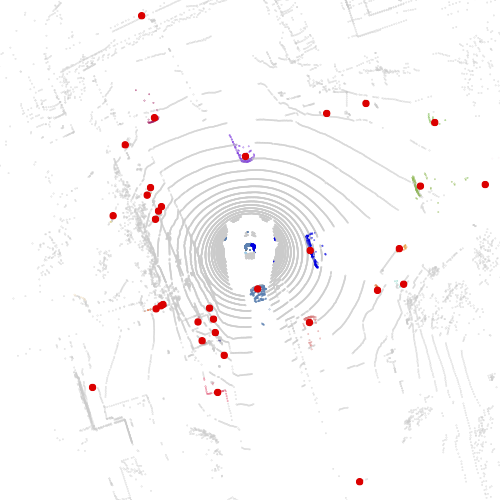}
        \caption{Instance Prediction}
    \end{subfigure}
    
\caption{Visualization of Panoptic-PolarNet on the nuScenes dataset. The red dots in the instance prediction represent the center for each instance.}
\label{fig:nuscenes_visualization}
\end{figure*}

We show the visualization examples of Panoptic-PolarNet on SemanticKITTI and nuScenes in Figure~\ref{fig:semkitti_visualization} and Figure~\ref{fig:nuscenes_visualization} respectively. Our method can make accurate instance predictions regardless of the distance and point density variation. We can also visually verify that nuScenes has significantly more instances than SemanticKITTI. And most of those instances belong to some challenging classes that have a small number of points. There are also duplicated instance predictions within a short distance. This could be fixed by introducing class-wise prior knowledge in the grouping stage in the future.

